%% file: acl_latex.tex
\pdfoutput=1

\documentclass[11pt]{article}

\usepackage[]{acl}
\usepackage{color}

\usepackage{times}
\usepackage{latexsym}

\usepackage[T1]{fontenc}

\usepackage[utf8]{inputenc}

\usepackage{microtype}
\usepackage{times}
\usepackage{CJKutf8}
\usepackage{xcolor}
\usepackage{latexsym}
\usepackage{tcolorbox}
\usepackage{multirow}
\usepackage{tikz}
\usepackage{capt-of}
\usepackage{graphicx}  
\usepackage{pgfplots}
\pgfplotsset{compat=1.12}
\usepackage{amsmath}
\usepackage{multicol}
\usepackage{booktabs}
\usepackage{color}
\usepackage{colortbl,array,xcolor}
\usepackage{xspace}

\newcommand{\zhtext}[1]{\begin{CJK*}{UTF8}{bsmi}{#1}\end{CJK*}}
\newcommand{\jatext}[1]{\begin{CJK}{UTF8}{min}{#1}\end{CJK}}
\newcommand{\kotext}[1]{\begin{CJK}{UTF8}{mj}{#1}\end{CJK}}
\newcommand{\data}{$\textsc{BUFFET}$\xspace}

\newcolumntype{R}[1]{>{\raggedleft\let\newline\\\arraybackslash\hspace{0pt}}m{#1}}
\usetikzlibrary{intersections}

\usepackage{caption}
\usepackage{subcaption}
\usepackage{graphicx}
\usepackage{amsfonts}
\usepackage{booktabs}
\usepackage{arydshln}
\usepackage{colortbl}
\usepackage{enumitem}
\usepackage{soul}
\usepackage{colortbl,array,xcolor}
\newcommand{\hlc}[2][yellow]{{%
    \colorlet{foo}{#1}%
    \sethlcolor{foo}\hl{#2}}%
}
\usepackage{color, colortbl}
\definecolor{azure}{rgb}{0.0, 0.5, 1.0}

\title{\data: Benchmarking Large Language Models \\for Few-shot Cross-lingual Transfer}

\author{\parbox{0.9\linewidth}{\centering{
Akari Asai$^{\dagger}$, Sneha Kudugunta$^{\ddagger}$, Xinyan Velocity Yu$^\dagger$, \\ Terra Blevins$^\dagger$, Hila Gonen$^\dagger$, 
Machel Reid$^\ddagger$, \\ Yulia Tsvetkov$^\dagger$, Sebastian Ruder$^\ddagger$, Hannaneh Hajishirzi$^{\dagger\heartsuit}$ \\
{\rm $^\dagger$University of Washington~~$^\ddagger$Google DeepMind ~~$^\heartsuit$Allen Institute for AI} \\
\url{https://buffetfs.github.io/} \\
}}
}

\begin{document}
\maketitle
\begin{abstract}
Despite remarkable advancements in few-shot generalization in natural language processing, most models are developed and evaluated primarily in English. 
To facilitate research on few-shot cross-lingual transfer, we introduce a new benchmark, called \data, which unifies 15 diverse tasks across 54 languages in a sequence-to-sequence format and provides a fixed set of few-shot examples and instructions. 
\data is designed to establish a rigorous and equitable evaluation framework for few-shot cross-lingual transfer across a broad range of tasks and languages. 
Using \data, we perform thorough evaluations of state-of-the-art multilingual large language models with different transfer methods, namely in-context learning and fine-tuning. 
Our findings reveal significant room for improvement in few-shot in-context cross-lingual transfer. In particular, ChatGPT with in-context learning often performs worse than much smaller mT5-base models fine-tuned on English task data and few-shot in-language examples. 
Our analysis suggests various avenues for future research in few-shot cross-lingual transfer, such as improved pretraining, understanding{, and future evaluations. }
\end{abstract}

\section{Introduction}
\input{sections/intro}
\section{Background and Related Work}
\input{sections/background}

\section{Benchmark: \data}
\input{sections/benchmark}


\section{Benchmarking LMs on \data}
\label{sec:evaluation}
\input{sections/experiments}

\section{Results and Analysis}
\input{sections/results}

\input{sections/analysis}


\section{Conclusion and Discussion}
\input{sections/discussion}

\section*{Limitations}
As the first step toward standardized evaluation for few-shot cross-lingual transfer, \data~focuses on popular discriminative tasks and some generative tasks. 
It does not include many datasets that require complex reasoning tasks, as noted above. 
Since our main focus is to benchmark different LLMs and learning methods in a comparable format, we do not explore sophisticated prompting methods, which can further boost performance. 
We anticipate that \data will encourage the LLM community to explore new methods to further improve in-context learning beyond English. 
We use instructions translated by the NLLB~\cite{costa2022no} for \textsc{Target ICL}; such machine-translated instructions are prone to errors, especially in less-represented languages, that can affect the final performance.

\section*{Ethics Statement}
{
While there has been significant research on in-context learning with LLMs, most of the focus has been on the English language. This raises questions about the applicability of findings from English few-shot NLP to few-shot cross-lingual transfer scenarios.
To address this gap, \data aims to provide a comprehensive and less biased evaluation framework.
However, it is important to note that our benchmark dataset currently covers only 57 out of the approximately 6,000 world languages. 
Moreover, we do not specifically focus on finer-grained language varieties and dialects that are commonly spoken by underrepresented populations.
In light of these limitations, we encourage future research to explore the effectiveness and limitations of widely-used transfer methods in a more diverse range of languages. 
This will help us gain a deeper understanding of the generalizability of transfer learning techniques across different linguistic contexts. 
}

\section*{Acknowledgements}
This research was supported by NSF IIS-2044660, ONR N00014-18-1-2826, ONR MURI N00014- 18-1-2670, DARPA MCS program through NIWC Pacific (N66001-19-2- 4031), and Allen Distinguished Award. AA is supported by the IBM fellowship. 
We are grateful to Orevaoghene Ahia for her help with ChatGPT evaluations. 
We thank our volunteer translators, Joongwon Kim, Usharani Injeti, and Sven Dorkenwald, for their help with translating instructions into different languages. 
Finally, we extend our appreciation to Jonathan H. Clark, Orevaoghene Ahia, Sandy Kaplan, and UW NLP researchers for their feedback on this draft.

\bibliography{abbr,custom}
\bibliographystyle{acl_natbib}
\clearpage
\appendix
\section*{Appendix}
\label{sec:appendix}
\input{sections/appendix}

\end{document}

%% file: sections/intro.tex
Recent advances in NLP primarily focus on the English language~\cite{blasi-etal-2022-systematic}. Due to the lack of sufficient training data in most of the world's languages~\cite{yu2022counting}, prior work explores direct transfer of pretrained language models to new languages after fine-tuning on resource-rich languages ({\it zero-shot cross-lingual transfer}; ~\citealt{XTREME}). 
Transferring after training a model on a few examples ({\it few-shot cross-lingual transfer}) often boosts performance, especially in languages that are distant from the source language \cite{lauscher-etal-2020-zero,hedderich-etal-2020-transfer}. 

{{In English, zero- or few-shot learning via in-context learning is an active area of research~\cite{beltagy-etal-2022-zero,schick-schutze-2021-exploiting,shin2020autoprompt}. 
In this learning paradigm, one prompts a large language model (LLM)  with few-shot demonstrations or natural language instructions to adapt to a new task, without any parameter updates. }
Yet, few-shot transfer across languages is still under-explored~\cite{xglm} in a wide range of tasks and languages. 
Moreover, it is unclear how effectively in-context learning performs in comparison to widely-used fine-tuning-based transfer methods under a comparable setup. 
}


\begin{figure}[t!]
\centering
\includegraphics[width=7.5cm]{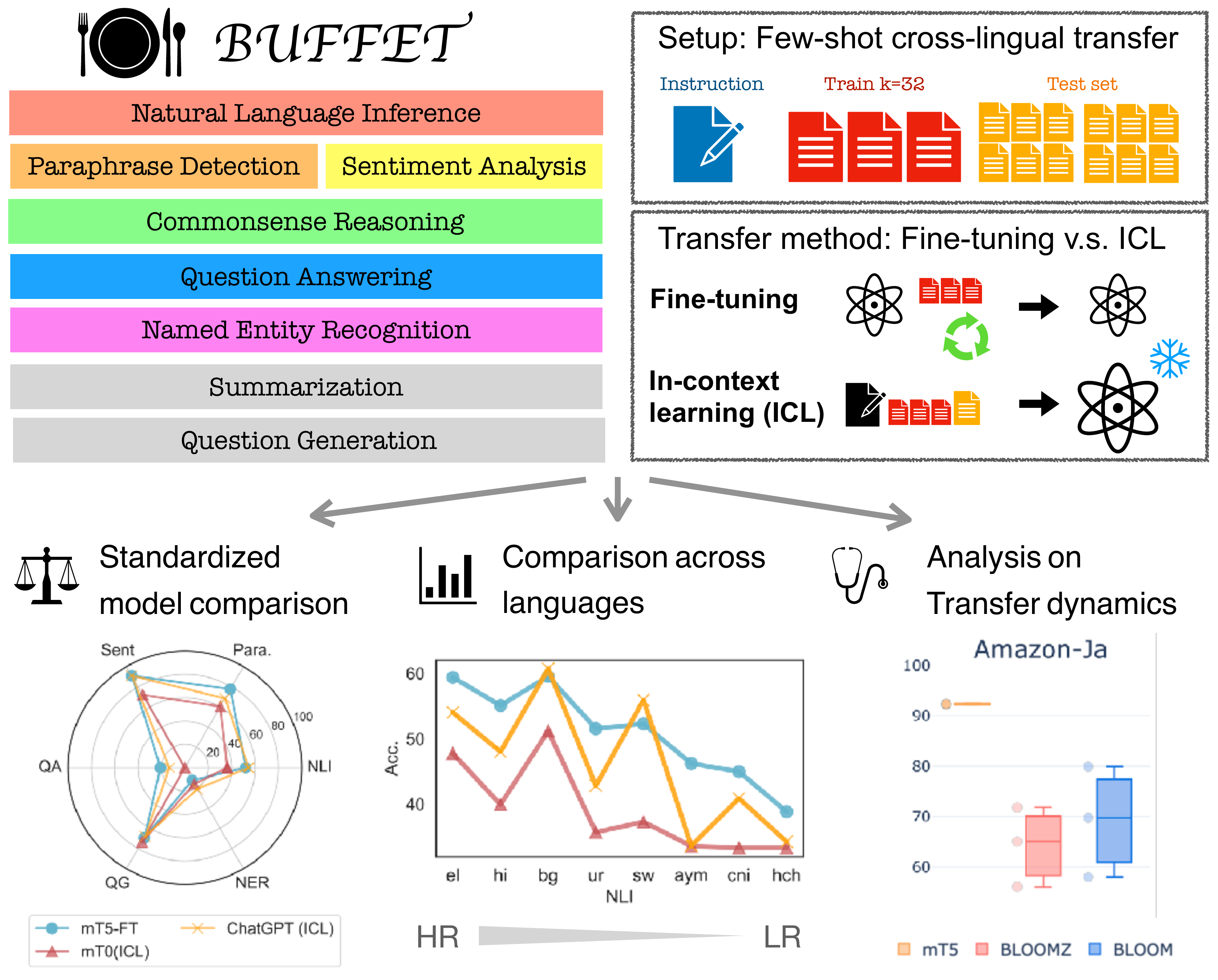}\caption{
\data includes unified diverse tasks in the same format, covering many typologically diverse languages. It enables a fair comparison across models, transfer methods, and languages and facilitates large-scale analysis across different learning setups. 
} 
\label{fig:teaser}
\end{figure}

This work introduces a new benchmark called \data: {\bf B}enchmark of {\bf U}nified {\bf F}ormat {\bf FE}w-shot {\bf T}ransfer Evaluation (Figure~\ref{fig:teaser}) to enable rigorous evaluations and advance research on few-shot cross-lingual transfer. 
Similar to a rich buffet, BUFFET curates a diverse mix of tasks: 15 different tasks---including classification, structured prediction, and natural language generation---across 54 languages. 
\data has several unique characteristics that are not present in prior multi-task multilingual benchmarks (summarized in Table~\ref{tab:comparison_benchmark}):    
{
\begin{itemize}[itemsep=-3pt,topsep=5pt,leftmargin=10pt]
\item \data provides a fixed set of few-shot examples for training and validation, allowing for fair comparisons across LMs and transfer methods. 
\item \data includes datasets annotated in each language or covering under-represented languages, which are often not included in existing multi-task benchmarks. 
\item \data combines diverse tasks into a unified text-to-text format and provides a set of English and machine-translated instructions for each task, removing the burdens of task-specific architecture changes or prompt engineering.  
\end{itemize}
}

Using this new benchmark, we extensively evaluate the current state-of-the-art multilingual large language models (LLMs), including mT5~\cite{xue2020mt5}, mT0~\cite{muennighoff2022crosslingual}, BLOOM~\cite{scao2022bloom}, BLOOMZ~\cite{muennighoff2022crosslingual}, and ChatGPT~\cite{ouyang2022training}, using both fine-tuning and in-context learning approaches. 
{In particular, \data enables us to investigate the following research questions:}

\noindent{{\bf (RQ1) Is in-context learning competitive with fine-tuning in few-shot cross-lingual transfer?}}
Notably, {given the same small numbers of examples in the target languages, in-context learning on LLMs (including ChatGPT, the most powerful model we evaluate in this work)  often under-performs much smaller specialized mT5-base models, as shown in Figure~\ref{fig:teaser} (bottom left). }

\noindent{{\bf (RQ2) How well do different transfer methods perform across tasks and languages? }}
{The performance gap between in-context learning-based baselines and fine-tuning-based baselines is more significant in under-represented languages (Figure~\ref{fig:teaser} bottom center). 
On NLI in indigenous languages of the Americas, ChatGPT or mT0-11B using in-context learning performs barely above random, while 580 million parameter mt5-base fine-tuned models retain strong performance. 
}
On the contrary, these LLMs perform well on generative tasks where a smaller task-specific model struggles, demonstrating their superiority in generating fluent text for diverse languages without abundant training data. 

\noindent{{\bf (RQ3) How does the choice of transfer setup affect different transfer strategies? }}
{\data~also enables us to perform an in-depth and extensive analysis of the effects of diverse demonstrations and instructions on the downstream transfer quality. }
Our observations indicate that the choice of few-shot training examples has a substantial influence on a model's performance, particularly, with greater variability in in-context learning, compared to fine-tuning.  
We note that optimal transfer settings may differ across models. For example, instruction-tuned models often face challenges in effectively utilizing few-shot samples and their performance deteriorates as the number of demonstrations increases, possibly because they are optimized for the zero-shot instruction-tuned training scheme. 
This highlights the need for a standardized benchmark to facilitate fair comparisons and further studies to assess such transfer dynamics in non-English data. 

Grounded in our analysis, we suggest avenues for future research in few-shot cross-lingual transfer for both dataset creation and model development. 
Our data and code are available online.\footnote{\url{https://buffetfs.github.io/}}

\begin{table}[t!]
\small
    \centering
    \begin{tabular}{l|ccccc }
\toprule
& Multi-ling. & Few-S & Gen. & Low-R  \\\midrule
XTREME & \textcolor{blue}{$\checkmark$}   &  &  \\
XTREME-R & \textcolor{blue}{\checkmark}  &  &  \\
XGLUE & \textcolor{blue}{\checkmark}  &  &   \textcolor{blue}{\checkmark}  \\
CrossFit &   & \textcolor{blue}{\checkmark} & \textcolor{blue}{\checkmark} & \\
MEGA* & \textcolor{blue}{\checkmark}  & \textbf{\textcolor{blue}{\checkmark}} & \\\hdashline
\data &   \textcolor{blue}{\checkmark}  & \textbf{\textcolor{blue}{\checkmark}} &\textbf{\textcolor{blue}{\checkmark}} & \textbf{\textcolor{blue}{\checkmark}}\\
\bottomrule
 \end{tabular}
    \caption{Comparison of the existing benchmarks based on their {multilinguality (Multi-ling.), few-shot task formulation (Few-S), availability of generative tasks (Gen.), and coverage of low-resource languages (Low-R)}.
    $^*$ indicates concurrent work. 
    }
    \label{tab:comparison_benchmark}
\end{table}

%% file: sections/background.tex
\subsection{Problem Formulation}
Due to the lack of annotated training data in many languages~\cite{blasi-etal-2022-systematic,yu2022counting,joshi-etal-2020-state}, transferring models trained on resource-rich languages (e.g., English) to other languages has been actively studied in multilingual NLP. 
In this paper, our main focus is on {\bf few-shot cross-lingual transfer}~\cite{lauscher-etal-2020-zero},  where a model is adapted using only a limited number of training or validation examples in the target language $L$. 
Another popular paradigm is {\bf zero-shot cross-lingual transfer}~\cite{artetxe-etal-2020-cross,XTREME} from English, where a model has access to training sets or instructions in English but not in the target language.

{
Various transfer methods have been investigated in the field, including the in-context learning methods (Section~\ref{sec:background_method}). Yet, limited research explores different transfer methods \emph{under comparable conditions}. 
With our new benchmark, \data, we facilitate fair comparisons between models and learning methods, establishing a basis for studying the dynamics of few-shot transfer across various languages (Section~\ref{sec:benchmark}).
}

\subsection{Benchmarks for Cross-lingual Transfer}
\label{sec:benchmark}
{To enable a scalable and rigorous evaluation across multiple tasks, prior work has proposed multi-task benchmarks that unify diverse existing datasets. 
XTREME~\cite{XTREME}, XTREME-R~\cite{ruder-etal-2021-xtreme} and XGLUE~\cite{liang-etal-2020-xglue} focus on zero-shot transfer of models fine-tuned on English datasets. 
Despite English-based few-shot evaluation benchmarks, such as CrossFit~\cite{ye-etal-2021-crossfit}, in few-shot cross-lingual transfer, we lack a standardized evaluation benchmark to facilitate the comparison of models and learning methods at scale. 
\data provides the first large-scale few-shot cross-lingual transfer suits to address the gap. 
Importantly, to mitigate the effects of the high-performance variance in few-shot cross-lingual transfer~\cite{zhao-etal-2021-closer}{, we curate and aggregate results from multiple fixed $k$-shot training instances for each task and language. }
Concurrent with our work, MEGA~\cite{ahuja2023mega} conducts experiments of few-shot cross-lingual transfer with a focus on classification and question answering tasks. 
\data unifies diverse tasks including both discriminative and generative tasks. 
We also include datasets covering languages underrepresented in prior work (e.g., African and indigenous languages). 
Table~\ref{tab:comparison_benchmark} summarizes the key differences between \data and prior benchmarks.  
}

\subsection{Methods for Cross-lingual Transfer}
\label{sec:background_method}
\paragraph{Fine-tuning-based approaches.}
Multilingual pre-trained models~\cite{devlin2018bert,xue2020mt5,conneau2019unsupervised} have the ability to adapt to new languages with no or few training instances in a target language~\cite{conneau-etal-2020-emerging,XTREME,wu-dredze-2019-beto}. 
\citet{lauscher-etal-2020-zero} and \citet{hedderich-etal-2020-transfer} report that particularly in languages that are distant from the source language, further fine-tuning model on few-shot samples greatly improves performance. 

\paragraph{Cross-lingual in-context learning.}
In-context learning~\cite{brown2020language} aims at making an LM learn a new task by conditioning on a task description (instruction) and training examples (demonstrations). 
Despite active research on in-context learning~\cite{schick-schutze-2021-just,min-etal-2022-rethinking}, most prior work focuses only on English. 
Recent work~\cite{xglm,muennighoff2022crosslingual} introduces pre-trained LMs trained on more multilingual pre-trained corpora or translated datasets and shows improved results. 
While prior evaluations often focus on classification or translation tasks~\cite{zhu2023multilingual,vilar2022prompting}, more recently~\citet{Shi2023language}, evaluate the use of instructions, demonstrations, and rationales in different languages across multiple reasoning tasks. 
However, how much LLMs with respect to in-context learning compete with the aforementioned fine-tuned approaches in a {\it comparable} setup and at scale has yet to be investigated, as they often use a large number of training examples in target languages~\cite{bang2023multitask}. 
{We demonstrate even with a small number of training examples, fine-tuning methods are competitive with in-context learning for cross-lingual transfer. }

%% file: sections/benchmark.tex
\begin{table*}[t!]
\small
    \centering
    \begin{tabular}{l| lllllll}
\toprule
Tasks & Dataset & Output & $|L|$ & $k$ &Metric & Domain & Data curation \\\midrule
\rowcolor{red!5}
NLI & \textsc{XNLI} & 3-way class & 14 & 16 & acc. & misc. &  translation \\
\rowcolor{red!5}
  & \textsc{Americas NLI} & 3-way class & 10 & 16 & acc. & misc. & translation \\ 
\rowcolor{red!5}
   & \textsc{Parsi NLU} & 3-way class & 1 & 16 &  acc. & misc. & in-language  \\ 
   \rowcolor{red!5}
       \rowcolor{red!5}
      & \textsc{OCNLI} & 3-way class & 1 & 16 &  acc. & misc. & in-language  \\ 
\rowcolor{red!5}
 & \textsc{KLUE-NLI} & 3-way class & 1 &  16 & acc. &  misc. & in-language  \\ 
 \rowcolor{orange!20}
Paraphrase Detection & \textsc{PAWS-X} &  2-way class & 6 & 7 & acc. & Wikipedia  & aligned \\
\rowcolor{yellow!20}
Sentiment & \textsc{Indic-NLU-sent.} &  2-way class & 14 & 16 & acc. & e-commerce & translation \\
\rowcolor{yellow!20}
Analysis & \textsc{Amazon Review} &  2-way class & 5 &16& acc. & e-commerce & in-language  \\
\rowcolor{green!10}
Commonsense & \textsc{XCOPA} & multi-choice &11  &  16 & acc. & misc. & translation \\
 \rowcolor{green!10}
 Reasoning & \textsc{XWinograd} & multi-choice &4  & 8 & acc. & misc. & translation \\
  \rowcolor{blue!10}
QA & \textsc{TyDiQA} & span & 8 & 8 & F1 & Wikipedia & in-language \\
  \rowcolor{purple!5}
Named Entity  & \textsc{WikiANN}& names \& tags &  33 & 32 &  F1 & Wikipedia & aligned \\
  \rowcolor{purple!5}
Recognition & \textsc{MasakhaNER}& names \& tags &  9  & 32 &  F1 & News & in-language \\\midrule
   \rowcolor{gray!10}
Summarization & \textsc{XLSUM} & summary & 12 & 1 & ROUGE &  News & aligned\\
   \rowcolor{gray!10}
 Question Generation  & \textsc{TyDi QA-QG} & question & 8 & 8 & BLEU  & Wikipedia & in-language  \\
\bottomrule
 \end{tabular}
    \caption{\textbf{The eight target tasks built upon 15 existing datasets in \data.} $|L|$ indicates the number of languages, and $k$ indicates the total number of training instances. 
    We include datasets that are diverse in terms of output format, tasks, and domains. We also include datasets that are curated by translation, in-language annotation (in-language) and automatically aligned (aligned) following \citet{yu2022counting}. 
    }
    \label{tab:task_overview}
\end{table*}
\label{sec:data}
We introduce a new standardized few-shot cross-lingual evaluation benchmark: \data ({\bf B}enchmark of {\bf U}nified {\bf F}ormat {\bf Fe}w-shot {\bf T}ransfer Evaluation). \data unifies diverse NLP tasks and provides fixed sets of few-shot samples per task to facilitate consistent comparisons (Table~\ref{tab:task_overview}).

\subsection{Design Principles}
We create the \data benchmark to establish a rigorous and equitable evaluation framework for few-shot cross-lingual transfer across a broad range of tasks and languages. We adhere to the following design principles with our benchmark.

\paragraph{Standardized few-shot samples. }
\data provides three different training and validation sets of $k$-shots (e.g., $k=32$) per task for a non-classification task, or per class for a classification task, for each language. 
\paragraph{Task diversity.}
Existing cross-lingual benchmarks often focus on classification or retrieval~\cite{XTREME,ruder-etal-2021-xtreme,liang-etal-2020-xglue}.
\data encompasses a broad range of task types, such as classification, generation, extraction, and structured prediction tasks. 
By converting all tasks into the same text-to-text format, we eliminate the need for task-specific model modifications or template conversions.  
\paragraph{Language diversity.}
{
\data~covers 54 typologically diverse languages, spanning 24 language families, including under-represented languages (e.g., indigenous languages of the Americas,  African languages). 
The 36 out of 54 languages are not Indo-European languages.  
A full list of languages is available in Appendix Table~\ref{tab:full_lang_list}. 
}

\paragraph{Beyond evaluations on translated data. }
Prior few- or zero-shot evaluations were often conducted on widely-used datasets translated from English (e.g., XNLI; \citealt{conneau-etal-2018-xnli}, XCOPA;~\citealt{ponti-etal-2020-xcopa}). 
Those datasets might exhibit undesired biases, such as translation artifacts or unnatural topic distributions~\cite{clark-etal-2020-tydi,artetxe-etal-2020-call}. 
We collect both translation-based datasets and datasets that are annotated directly in each language (Table~\ref{tab:task_overview}, Data curation).

\subsection{\data Construction Process}
\begin{figure}[t!]
\centering
\includegraphics[width=7.5cm]{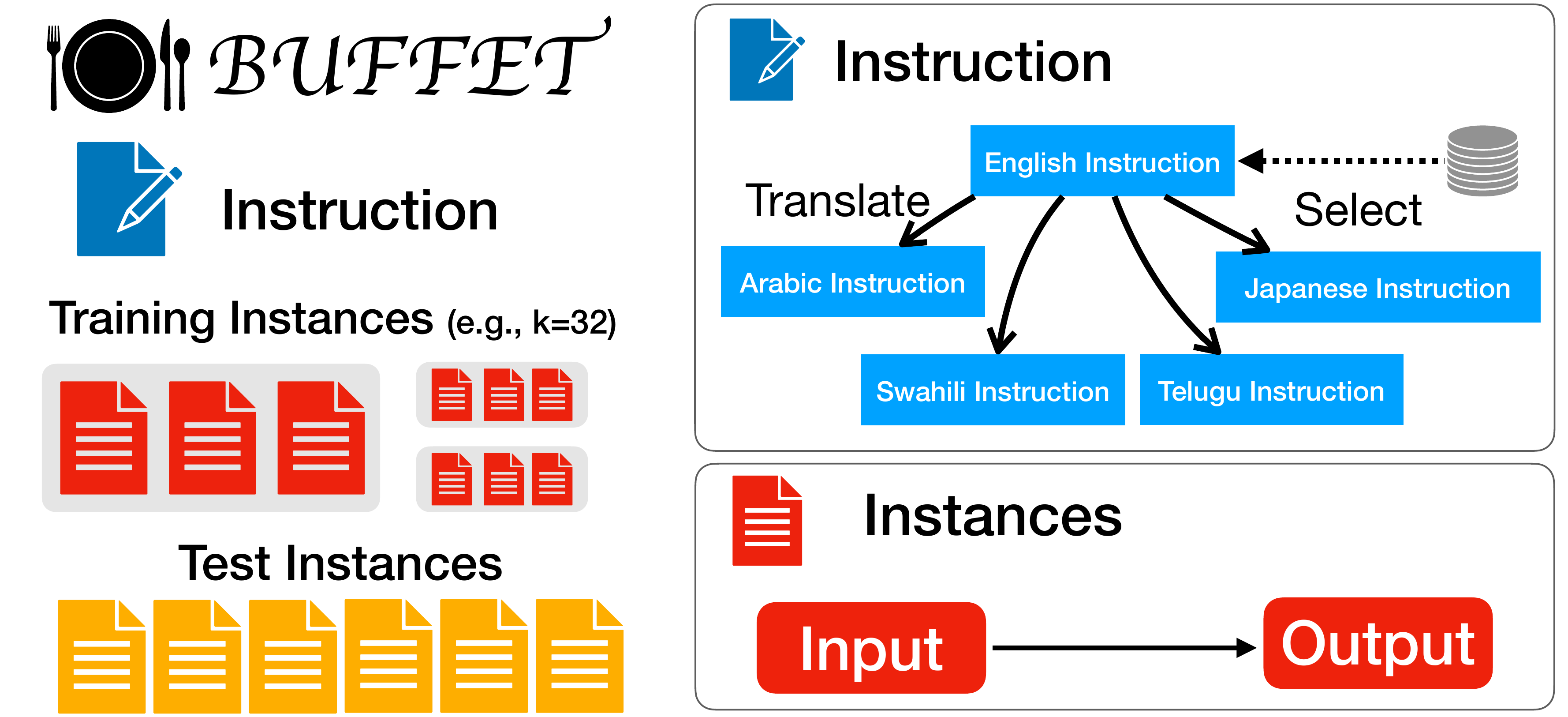}\caption{
\data includes 15 datasets, which are unified into the same single text-to-text format. 
} 
\label{fig:buffet_example}
\end{figure}

{
Following \citet{ye-etal-2021-crossfit}, we unify all tasks into the same text-to-text format, where a model is expected to directly generate the desired outputs given diverse inputs~\cite{roberts2019exploring}. 
For each dataset in \data, we unify instance representations of {\it instruction}, $k$-shot {\it instances} for training and validation. 
Each training instance consists of an input and output. 
Figure~\ref{fig:buffet_example} shows an overview. 
Section~\ref{sec:construction_overview} provides the outline of the unification, and Section~\ref{sec:task_details} provides a task-specific process. 
}

\subsubsection{Unification Process}
\label{sec:construction_overview}
\paragraph{Few-shot instance selection.}
By default, we use all of the languages included in the original datasets. For automatically aligned datasets with many test languages, such as XLSUM or \textsc{WikiANN}, we filter out languages that are not included in any other \data datasets following suggestions by \citep{yu2022counting}.\footnote{
On \textsc{XLSUM}, we further reduce the number of languages to reduce the inference costs while maintaining language diversities. }
{For each language in each dataset, we randomly sample $k$-shot instances (or {\it demonstrations}) for training and validation sets using the same random seeds.\footnote{We use 100, 13, and 21 as seed numbers, following \citet{ye-etal-2021-crossfit}. 
Once we sample the instances, we fix the training and validation sets. }
With large-scale automatically aligned datasets, we randomly sample 1,000 test instances in XLSUM and \textsc{WikiANN} and 2,000 test instances for Amazon Review, to reduce inference time costs across many languages and multiple sets of demonstrations. }

\paragraph{Instruction selection.}
We use English instructions from SuperNaturalInstructions~\cite{wang-etal-2022-super} and PromptSource~\cite{bach2022promptsource}.
Among multiple annotated instructions, we sample the first instruction for a similar task that suits our text-to-text scheme. 
For some tasks, we modify the original instruction to make labels consistent with the names used in \data\footnote{For example, an instruction for PAWS-X says the class names are ``repeated/not repeated'' while in \data we use ``duplicated/not\_duplicated'' as labels, so we change the labels in the original instruction.} or to remove task-specific dependencies in the input data field. See Appendix Table~\ref{tab:task_example} for the full list of instructions.

\paragraph{Instruction translation.}
Despite rapid progress of instruction-tuning in English LLMs~\cite{wei2021finetuned,sanh2022multitask, mishra-etal-2022-cross, wang-etal-2022-super}, cross-lingual setups still lag behind due to a lack of instructions in the target languages. Prior work often translates instructions for the target tasks \cite{xglm,Shi2023language}. 
We provide translated instructions for 15 datasets in 54 target languages, translated by NLLB~\cite{costa2022no}, and manually translate the instructions into five languages.\footnote{Manual translations are performed by bilingual volunteers. }

\subsubsection{Tasks and Dataset Curation}
\label{sec:task_details}
{
We first select eight popular NLP tasks and, for each task, we identify available datasets using a careful survey of multilingual datasets by \citet{yu2022counting}.
}
{Appendix} Table~\ref{tab:task_example} shows examples. 
\paragraph{{Natural language inference}.}
{Natural Language Inference (NLI) involves determining the logical relationship (i.e., \texttt{entailment}, \texttt{contradiction}, \texttt{neutral}) between two text fragments, i.e., a premise and a hypothesis. 
In addition to the widely used \textsc{XNLI}~\cite{conneau-etal-2018-xnli}, we gather NLI datasets that are annotated in each language or designed to cover extremely under-represented languages: \textsc{AmericasNLI}~\cite{ebrahimi-etal-2022-americasnli}, \textsc{ParsiNLU-Entailment}~\cite{10.1162/tacl_a_00419}, \textsc{KLUE-NLI}~\cite{park2021klue}, and \textsc{OCNLI}~\cite{hu-etal-2020-ocnli}.}
We use 16 examples for each class.
\paragraph{Paraphrase detection.}
Paraphrase detection is the task of identifying whether two sentences have/do not have the same meaning (\texttt{duplicate} or \texttt{not duplicated}). 
We adopt \textsc{PAWS-X}~\cite{yang-etal-2019-paws} and include 16 shots for each class as few-shot training and validation data. 
\paragraph{Sentiment analysis.}
Binary sentiment analysis identifies whether a text (e.g., a product review from Amazon) expresses \texttt{positive} or \texttt{negative} sentiment towards a topic. We use the \textsc{Multilingual Amazon Review dataset}~\cite{keung-etal-2020-multilingual} and \textsc{IndicNLU-Sentiment}~\cite{aggarwal-etal-2022-indicxnli}. For the former, we discard the neutral class (the reviews with a score of 3) and assign reviews with scores of 4 and 5 to the positive class and reviews with scores of 1 and 2 to the negative class. For both datasets, we sample 16 demonstrations per class. 

\paragraph{Commonsense reasoning.}
We use two commonsense reasoning datasets, \textsc{XCOPA}~\cite{ponti-etal-2020-xcopa} and \textsc{XWinograd} \cite{muennighoff2022crosslingual}.
Given a sentence and two options, a model selects one of the option labels, \texttt{(A)} or \texttt{(B)}, based on which is better suited to the given context. 
Due to the smaller scale of the datasets, we sample 16 and 8 training instances in total for \textsc{XCOPA} and \textsc{XWinograd}, respectively. 

\paragraph{Question answering.}
Question Answering (QA) is the task of answering a question given a paragraph, where the answer is a sub-span of the paragraph. We use \textsc{TyDiQA-GoldP}~\cite{clark-etal-2020-tydi}, which we refer to as \textsc{TyDiQA} for simplicity. 
Due to the longer average input length, we limit the number of exemplars to 8. 

\paragraph{Named entity recognition.}
Named Entity Recognition (NER) is a representative sequence labeling task, where a system detects and classifies named entities in an input sentence. We adopt \textsc{WikiANN}~\cite{pan-etal-2017-cross} and \textsc{MasakhaNER}~\cite{10.1162/tacl_a_00416}. 
Though \textsc{WikiANN} covers 216 languages, we exclude languages that are covered only by \textsc{WikiANN} or XLSUM due to the aforementioned issues. 
We convert the task into a text-to-text format, where given an input sentence, a model extracts all named entities with named entity tags:\footnote{{This is more challenging than the standard sequence labeling setup since the model must reproduce the entity spans and generate appropriate tags. For example, the output for ``Obama served as the 44th president of the United States.'' would be ``Obama <person> United States <location>.''}} \texttt{<location>}, \texttt{<person>}, and \texttt{<organization>}.\footnote{Although \textsc{MasakhaNER} supports other named entity tags and distinguishes the beginning and middle/end of the named entities, we discard named entity categories beyond the three types and merge the beginning and middle/end entity tags to make the task formulation consistent with \textsc{WikiANN}.}
We use 32 instances overall for few-shot transfer.

\paragraph{Summarization.}
We use the \textsc{XLSum}~\cite{hasan-etal-2021-xl} dataset to benchmark models' ability to generate a summary given a news article. Due to the context window limit, we use only 1 shot for training in this task. 

\paragraph{Question generation.}
Question generation generates a question according to a given input passage and a corresponding answer~\cite{10.5555/3491440.3491993}. 
We convert the \textsc{TyDiQA-GoldP} dataset into a question generation task{, which we refer to \textsc{TyDiQA-QG}. }
Given the gold paragraph and an answer, the system generates the original question. 
We use 8 examples for few-shot training.

\subsection{\data Evaluation}
\subsubsection{Evaluation Metrics}
Table~\ref{tab:task_overview} (Metric) lists task-specific metrics.
To mitigate the variance from different few-shot samples, for each language included in each task, we take the average of a model's performance given {three} different sets of $k$-shot instances.  
Subsequently, each dataset score is calculated as a macro-average of the per-language score~\cite{clark-etal-2020-tydi}. 
Finally, following \citet{liang-etal-2020-xglue}, we take two separate average scores: (a) {\bf Avg. class} score of all classification and QA tasks, and (b) {\bf Avg. generation} score of all generation tasks. 

\subsubsection{\data-Light}

Conducting a comprehensive evaluation covering a wide range of languages and tasks in \data, while undoubtedly necessary, can be a time-consuming process. We introduce \data-light, which contains a representative subset of languages and tasks for a rapid assessment even in resource-limited scenarios.
We carefully select languages and datasets to ensure that we cover a diverse range of languages and output formats, assuming limited resources.
See the overview of \data-light in Appendix Section~\ref{sec:buffet_light}. 

%% file: sections/experiments.tex
\begin{table}[t!]
\footnotesize
    \centering
    \begin{tabular}{lcccc }
\toprule
&\multicolumn{2}{c}{Training Demos} &  \multicolumn{2}{c}{Instructions}   \\
{\bf Transfer}  & EN & Target & EN & Target \\\midrule
\textsc{Target FT} & & \textcolor{red}{$k$} & \cellcolor{gray!30} & \cellcolor{gray!30} \\ 
\textsc{English FT} & \textcolor{olive}{$N$} &  & \cellcolor{gray!30} & \cellcolor{gray!30} \\
\textsc{Eng.+Tgt. FT} & \textcolor{olive}{$N$} & \textcolor{red}{$k$} & \cellcolor{gray!30} & \cellcolor{gray!30} \\ \midrule
\textsc{English ICL} & & \textcolor{red}{$k$} & \textcolor{blue}{\checkmark}   &  \\ 
\textsc{Target ICL} & & \textcolor{red}{$k$} & & \textcolor{blue}{\checkmark}  \\
\textsc{Z-EICL} & & & \textcolor{blue}{\checkmark} & 
\\
\midrule \midrule
{\bf Transfer} & { Pretraining} & \multicolumn{3}{r}{LMs} \\ \midrule
\textsc{Fine-tuning} & Unlabeled & \multicolumn{3}{r}{mT5-base} \\
\textsc{In-c. Learning} & Unlabeled & \multicolumn{3}{r}{BLOOM, mT5-xxl} \\
  \rowcolor{blue!10}
\textsc{In-c. Learning} & + Instruction &\multicolumn{3}{r}{BLOOM-7B, mT5-xxl} \\
  \rowcolor{blue!10}
 & \multicolumn{1}{l}{~~~Tuning} &\multicolumn{3}{r}{ChatGPT} \\
\bottomrule
 \end{tabular}
\caption{\textbf{Comparison of different few-shot and zero-shot transfer methods, based on the resources they use.} The top section requires parameter updates via fine-tuning (FT), and the bottom uses ICL with no updates. \textcolor{red}{$k$} = k-shot examples; \textcolor{olive}{$N$} = full training data; \textcolor{blue}{\checkmark}= instruction language. 
The bottom half lists the models evaluated in this work. \hlc[blue!10]{The blue-colored rows} are instruction-tuned models.   }
    \label{tab:method}
\end{table}

\subsection{Transfer Methods}
In this study, we investigate various transfer methods with and without parameter updates. 
To assess the benefit of $k$-shot training examples in the target language, we also conduct experiments on zero-shot transfer methods. 
We assume that the model can optionally use instructions in the target language or another language, or full training sets in a high-resource language like English. 
This assumption is reasonable given the abundance of labeled datasets in high-resource languages~\cite{yu2022counting,joshi-etal-2020-state} and the cheaper costs of instruction annotations. 
Table~\ref{tab:method} provides an overview of  different approaches, categorized according to the optional inputs they use during training or inference. 

\paragraph{Fine-tuning (methods with parameter updates).}
We explore several transfer approaches that require parameter updates. 
\begin{itemize}[itemsep=-3pt,topsep=5pt,leftmargin=10pt]
    \item {\bf Target fine-tuning (\textsc{Target FT})} trains models on few-shot samples for each language.
    \item {\bf English fine-tuning (\textsc{English FT})} trains models on a source language (i.e., English) only and uses no target language data. 
    \item {\bf English+Target fine-tuning (\textsc{Eng.+Tgt. FT})} first trains models on large-scale English datasets and then fine-tunes models on few-shot samples of target languages. 
\end{itemize}

\paragraph{In-context learning (methods without updates). }
We explore several in-context learning methods.
\begin{itemize}[itemsep=-3pt,topsep=5pt,leftmargin=10pt]
\item {\bf English in-context learning (\textsc{English ICL})} uses English instructions and demonstrations in the target languages.
\item {\bf Target ICL (\textsc{Target ICL})} uses both instructions and demonstrations in the target language. 
\item {\bf Zero-shot English In-context learning (\textsc{Z-EICL})} uses only English instructions without demonstrations (neither in English nor in the target language), as in zero-shot transfer.
\end{itemize}
Unlike in English, where abundant instructions and instance annotations are available, for many languages we often lack annotated instructions~\cite{wang-etal-2022-super}. 
We use machine-translated instructions in \data as the main baseline.

\subsection{Language Models}
{A key aspect of language models is their pretraining strategies. 
In addition to conventional pretraining using unlabeled corpora~\cite{devlin2018bert,brown2020language}, instruction-tuning has been actively studied; this approach trains an LLM on a massive number of tasks with instructions~\cite{muennighoff2022crosslingual,ouyang2022training,wei2021finetuned}.  
In this work, we evaluate six diverse models pretrained with different strategies~{ (Table~\ref{tab:method})}.  
}

\paragraph{Models for fine-tuning.}
Due to the high costs of fine-tuning for every $k$-shot setting, we experiment with an efficient yet competitive mT5-base with 580 million parameters~\cite{xue2020mt5}.

\paragraph{Models for in-context learning. } 
We experiment with BLOOM-7B (7 billion parameters; \citealp{scao2022bloom}) and mT5-xxl (13 billion parameters; \citealp{xue2020mt5}). 
We also experiment with their instruction-tuned variants: BLOOMZ-7B and mT0-xxl~\cite{muennighoff2022crosslingual}, as well as the current state-of-the-art ChatGPT (\texttt{gpt-3.5-turbo}; \citealt{ouyang2022training}). 
Note that these models are trained on some of the datasets included in \data. 
We do not exclude such overlapping datasets, but we indicate such seen tasks with $^*$ in the main result table.\footnote{It is unclear which datasets ChatGPT is trained on.}

\subsection{Experiment Details}
\paragraph{Fine-tuning.}
In all settings, we fine-tune models on few-shot samples for 300 epochs for \textsc{Target FT} and 200 epochs for \textsc{Eng.+Tgt. FT}. 
When fine-tuning LMs on large-scale English datasets (for both \textsc{Eng.+Tgt. FT} and \textsc{English FT}), we train for three epochs.
We use representative English datasets following \citet{XTREME}: \textsc{SQuAD}~\cite{rajpurkar-etal-2016-squad} for QA, \textsc{MNLI}~\cite{williams2017broad} for NLI, \textsc{PAWS}~\cite{zhang2019paws} for paraphrase detection, \textsc{XLSUM}~\cite{hasan-etal-2021-xl} for summarization, \textsc{COPA}~\cite{211227} for \textsc{XCOPA}, \textsc{Winograd} for \textsc{XWinograd}, the \textsc{Amazon Multilingual Review} English set for sentiment analysis, and the \textsc{TyDiQA-QG} English set for question generation.

\paragraph{In-context learning.}
We prompt LLMs with instructions and $k$-shot demonstrations available in \data. 
Different models have different maximum context window sizes: mT0 only accepts up to 1024 tokens, while BLOOMZ and ChatGPT accept up to 2048 and 4096, respectively.
We add training instances up to the maximum token length for each model and discard instances that do not fit the context window. 
{We found that mT0 often performs well-given zero or smaller numbers of few-shot samples. We use 4-shots for mT0~\textsc{English ICL} and \textsc{Target ICL}  by default.} 
We use greedy decoding for predictions. 
{
For tasks with a fixed set of pre-specified answer candidates, we compute the probability of option tokens by iterating all options except for ChatGPT without access to token probabilities. 
Due to the high inference costs, we evaluate ChatGPT only on \data-Light, 
}

\begin{table*}[t!]
\small
    \centering
    \renewcommand{\arraystretch}{1.3}
    \begin{tabular}{ l|r|ccccccccccc}
\toprule
  &  Output &  \multicolumn{3}{c}{Classification} &  \multicolumn{2}{c}{Multiple Choice} & Span & Str. &  \multicolumn{2}{c}{Generation} &  \multicolumn{2}{c}{Avg.}  \\

&Tasks & NLI& Sent. & PWX&XCPA&XWGD &  TyDi & NER & QG& Summ. & class & gen \\\midrule
Random &  & 33.3 & 50.0 & 50.0 & 50.0&  50.0 & -- & -- & -- & -- & -- & \\\hline
\textsc{Tgt. FT} & {mT5} & 35.0 &                67.2 &                  47.7 &   44.1 &       48.8 &                 5.2 &                    33.4 &                  3.2 &            2.5 &              40.7 &             2.9  \\
\textsc{Eng. FT} &  {mT5} & 49.9 &                89.8 &                  77.5 &    0.0 &        0.0 &                66.8 &                    39.0 &                  3.8 &            6.2 &              55.5 &             5.0 \\ 
\textsc{Eng.+Tgt.} & mT5 &   {\bf 51.8} &             {\bf 91.0} &                  {\bf 77.8} &   49.5 &       48.5 &               {\bf 69.5} &                   {\bf  47.8} &                 12.5 &          {\bf  11.8} &            {\bf 61.2} &            {\bf 12.2}   \\
\midrule
\textsc{Eng. ICL}& BLOOM&  32.1 &                81.7 &                  42.2 &   50.2 &       51.0 &                54.7 &                    24.2 &                  9.3 &            3.4 &              45.0 &             6.4   \\
& mT5&   35.7 &                50.0 &                  42.2 &   50.4 &       47.5 &                 0.2 &                     0.0 &                  0.0 &            0.4 &              31.7 &             0.2 \\
  \rowcolor{blue!10}
& BLOOMZ&  31.5 &                86.3* &                  48.5* &   50.4 &       54.2 &                65.8* &                    25.5 &                 13.5 &            8.3* &              47.5 &            10.9 \\
  \rowcolor{blue!10}
& mT0 &   36.2 &                72.1* &                  60.6* &   50.5 &       60.3 &                73.6* &                     7.9 &                 16.1 &            3.4* &              46.3 &             9.7  \\
  \rowcolor{blue!10}
 & ChatGPT$\dagger$ &   {\bf {54.5}} &                91.1 &                  68.6&  {\bf { 76.7 }} &       73.3 &                68.1 &                  45.4  &                 {\bf {21.2}} &            5.4 &              {\bf {64.6}} &            13.3 \\ 
\hdashline
\textsc{Tgt. ICL}& BLOOM &   27.9 &                80.5 &                  46.5 &   49.9 &       51.8 &                11.8 &                    23.4 &                 11.2 &            3.6 &              40.4 &             7.4 \\
& mT5& 35.7 &                50.0 &                  42.2 &   49.8 &       45.2 &                 0.2 &                     0.0 &                  0.0 &            0.4 &              31.5 &             0.2  \\
  \rowcolor{blue!10}
& BLOOMZ &   32.0 &                61.7* &                  52.5* &   49.7 &       55.5 &                63.1* &                    23.4 &                  9.1 &            8.0* &              43.4 &             8.5 \\
  \rowcolor{blue!10}
& mT0&  36.2 &                72.1* &                  60.6* &   50.5 &       60.3 &                73.6* &                     7.9 &                {\bf  16.1} &            3.4* &              46.3 &             9.7  \\
  \rowcolor{blue!10}
& ChatGPT$\dagger$ &48.2 &               {\bf { 91.5}} &                  68.2 &   74.3 &       {\bf {73.4}} &                68.0 &                    44.8 &                 21.1 &           {11.4} &              62.7 &           {\bf {  16.3}} \\
\hdashline
\textsc{Z-EICL}& BLOOM&    33.3  &       37.2  &         42.3 &   50.0 &     47.1 &               4.3 &                     0.0 &                  {14.0} &          6.3 &              29.2 &            10.1 \\
& mT5& 35.1  &               49.8 &                  42.2 &      {50.7}   & {55.5} &              2.2 &                     0.0 &                0.1 &            4.8 &              32.5 &            0.6  \\
  \rowcolor{blue!10}
& BLOOMZ & 33.5 &                51.6* &                  57.8* &   51.8 &       51.0 &                83.2* &                    11.2 &                  9.5 &            4.3* &              41.9 &             6.9 \\
  \rowcolor{blue!10}
& mT0& 48.5 &                90.0* &                  {{90.6*}} &  {\bf 63.8} &      {\bf 61.0} &              {{    80.1* }}&                     0.0 &                 10.2 &           {{12.0*}} &              56.4 &            11.1\\
\bottomrule
 \end{tabular}
    \caption{\textbf{Overall experiment results in \data}. Note that to enable comparison between ChatGPT (only tested on \data-Light) and other methods, we present \data-Light results, and the overall results on \data are available in Table~\ref{tab:main_results}. 
    {The blue-colored rows are instruction-tuned models, and we added $^*$ symbols next to the scores for the tasks on which the models have been trained. 
    ``Random'' shows random baseline performance.
    {\bf Bold} fonts indicate the best results for each task, among the models that are not directly trained on the task.  
    When ChatGPT achieves the best results, we also note the second-best number from the models that are not trained on the task, acknowledging the possibility that ChatGPT may have encountered a similar task during training.
    }}
    \label{tab:main_results_light}
\end{table*}

%% file: sections/results.tex
\subsection{Main Results}
\label{sec:main_results}
Table~\ref{tab:main_results_light} shows aggregated results  of fine-tuned and in-context learning-based LMs on \data.  
We show full experiment results on each task in the Appendix. 
Below, we summarize the key findings.

\paragraph{{LLMs with in-context learning often lag behind much smaller fine-tuned models. } } 
{
While in-context learning has shown remarkable performance in English, our comparison shows that few-shot cross-lingual transfer via in-context learning remains challenging; \textsc{English ICL} using {BLOOM, BLOOMZ (7 billion) and mT0 (13 billion) often under-perform mt5-base (580 million) fine-tuned on English datasets (\textsc{English FT} or \textsc{Eng.+Tgt. FT})}. 
}
However, when abundant English task data is not available, mT5-based fine-tuning methods (\textsc{Target FT}, or \textsc{Eng.+Tgt. FT} on \textsc{XCOPA} and \textsc{XWinograd}) often perform poorly and are outperformed by \textsc{English ICL} or \textsc{Target ICL} baselines.  
This implies that when lacking task-specific training data, prompting LLMs can be more effective.  

\paragraph{{Instruction-tuning helps in zero-shot but may not generalize for few-shot settings.}}
Table~\ref{tab:main_results} demonstrates that the zero-shot performance of instruction-tuned models is significantly higher than the zero-shot performance of non-instruction-tuned models: 
On average, both mT0-xxl and BLOOMZ-7B \textsc{Z-EICL}, demonstrate significantly better performance compared to their non-instruction tuned counterparts, namely mT5-xxl and BLOOM-7B \textsc{Z-EICL}, with margins of 12.7 and 23.9 points in Avg. class, respectively. 
It is worth noting that while the performance improvements on seen tasks contribute to these gains (indicated by *), mT0-xxl \textsc{Z-EICL} exhibits substantial advancements on unfamiliar tasks.
This further confirms the effectiveness of instruction-tuning in zero-shot transfer, as discussed in prior studies~\cite{muennighoff2022crosslingual,wei2021finetuned,mishra-etal-2022-cross}. 

{
However, our study also highlights a surprising performance deterioration when moving from zero-shot to few-shot settings for instruction-tuned models:  
across tasks, mT0 performs worse in few-shot settings than in zero-shot settings (\textsc{English ICL} v.s. \textsc{Z EICL}).
BLOOMZ shows performance gains from few-shot demonstrations; BLOOMZ \textsc{E ICL} achieves 44.3, outperforming BLOOMZ \textsc{Z EICL} by 5 points in Avg. class score. Yet, it also exhibits large performance declines on the tasks that are used during their instruction-tuning (\textsc{TyDiQA}, \textsc{PAWS-X}). 
Our hypothesis is that such instruction-tuned models are optimized to execute a new task solely based on an instruction, with no prior demonstrations~\cite{muennighoff2022crosslingual}, and may struggle to learn in context from few-shot demonstrations. 
We conduct controlled experiments in Section~\ref{sec:analysis} for further analysis. 
}

\paragraph{{Zero- or few-shot transfer remains challenging in under-represented languages.} }

\begin{figure*}[t!]
\centering
\includegraphics[width=15.5cm]{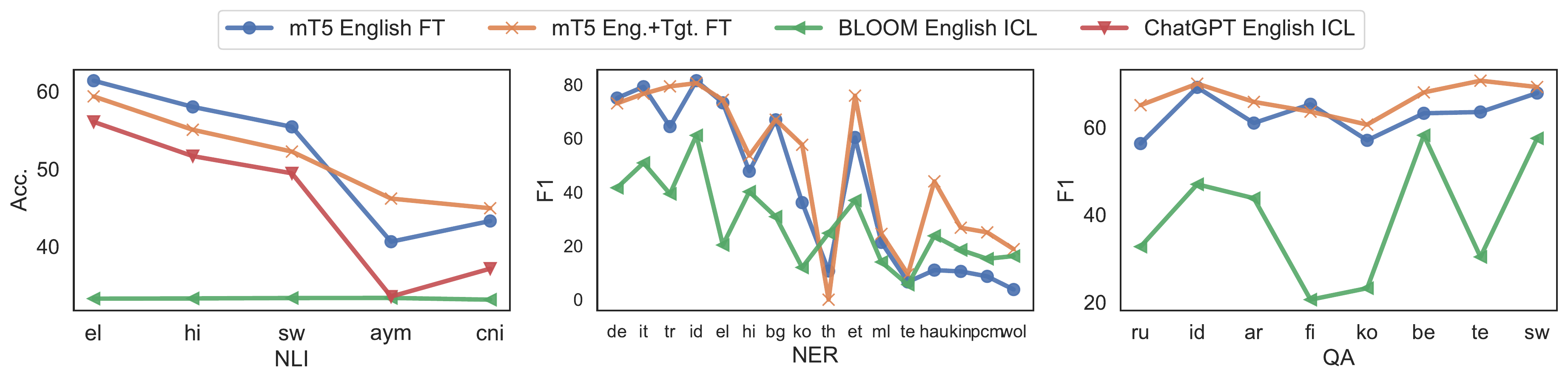}\caption{
\textbf{Model performance across three tasks, NLI, NER, and QA, displayed for various languages.} The languages are sorted based on token availability in mC4, with the left side representing high-resource languages. {All methods show performance deteriorations in lower-resource languages (right side), with larger drops in \textsc{English-ICL} methods. Additional fine-tuning in target languages is more effective in less-represented languages. }
} \label{fig:qa_lang}
\end{figure*}
Figure~\ref{fig:qa_lang} illustrates the performance of models on NER (\textsc{WikiANN} and \textsc{MasakhaNER}), NLI (\textsc{XNLI}, \textsc{AmericasNLI}), and QA (\textsc{TyDiQA}) tasks across different languages.
{
The languages are sorted based on the token availability in the mC4 corpus,\footnote{We use the token count statistics available at \url{https://github.com/allenai/allennlp/discussions/5265}. For languages that are not included during pretraining, we sort the language names alphabetically. } with high-resource languages positioned on the left side.
}
Our results indicate that the zero- or few-shot transferability of the model is often constrained in understudied languages.  
{
In NER and NLI tasks, a noticeable decrease occurs in performance from high-resource to low-resource languages. 
It's important to note that several languages included in \textsc{MasakhaNER} or \textsc{Americas NLI} are not part of the pretraining process. 
Models such as mT5 \textsc{English FT} or ChatGPT \textsc{English ICL}  exhibit strong performance in high-resource languages. 
However, their performance significantly drops in less-represented languages.  
For instance in Aymara (\texttt{aym}), ChatGPT achieves slightly higher performance than a random baseline, outperformed by mT5 \textsc{Eng.+Tgt. FT} by 13\%.  
 mT5 \textsc{Eng.+Tgt. FT} also significantly outperforms mT5 \textsc{English FT} in lower-resource languages, as indicated by the performance gap between the orange and blue lines in Figure~\ref{fig:qa_lang}. 
Notably,  mT5 \textsc{Eng.+Tgt. FT} outperforms  mT5 \textsc{English FT} by 30\% in Hausa on MasakhaNER. 
This indicates that fine-tuning with only $k$ instances in target languages can still greatly helps in less-represented languages. }

{
We also observe performance drops in Finnish, Korean, and Russian for BLOOM and BLOOMZ in \textsc{TyDiQA}. Finnish, Korean, and Russian are excluded from BLOOM pretraining,\footnote{\url{https://huggingface.co/bigscience/bloom}} which we attribute to these performance drops.  Conversely, mT5 fine-tuning-based methods consistently display strong performance across languages. Interestingly, in Bengali, which is often considered less represented, BLOOMZ achieves performance comparable to mT5 fine-tuned models. We also observe the same trends in BLOOMZ. These results suggest pretraining setup may strongly affect downstream task performance even after instruction tuning. 
}

\paragraph{ChatGPT has strong generation capabilities but requires careful instruction design.}
As discussed, though ChatGPT significantly outperforms other LLMs with in-context learning, its performance often lags behind fine-tuning-based methods in some discriminative tasks, particularly in less-represented languages.  
ChatGPT, however, significantly outperforms fine-tuned models on tasks that require target language generations (e.g., question generation, QA) with the exception of summarization (\textsc{XLSUM}). 
On \textsc{XLSUM}, we found that ChatGPT often generates semantically correct summarizations in English rather than in the input article language, resulting in low ROUGE-2 scores. 
We do not observe that phenomenon in other LLMs (e.g., BLOOMZ); we show some ChatGPT output examples in the Appendix Table~\ref{tab:xlsum_examples}. 
Though more prompt engineering can boost ChatGPT's performance in summarization~\cite{huang2023prompting}, we use the same prompts throughout the evaluations for a fair comparison. 
We also observe that when instructions are given in the target language, ChatGPT often outputs a summary in the language, as shown in improved \textsc{XLSUM} performance in ChatGPT ~\textsc{Target ICL}. 

\begin{figure*}[t!]
\centering
\includegraphics[width=15.5cm]{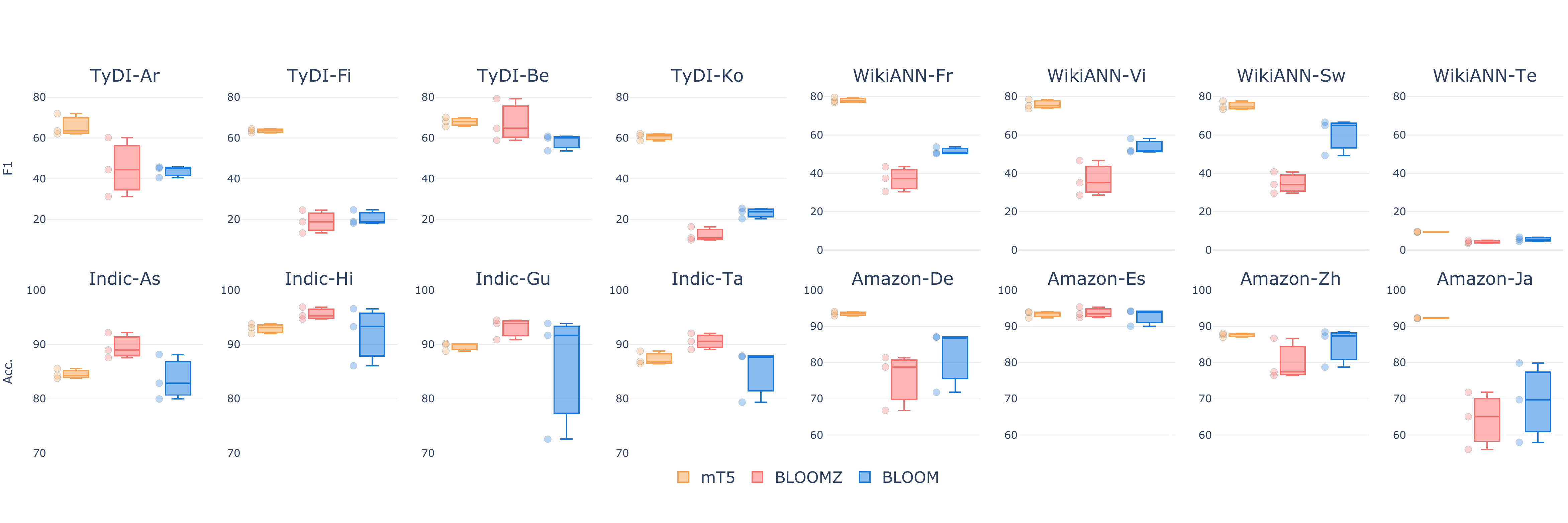}\caption{
\textbf{Model performance across different $k$-shot demonstrations for QA (\textsc{TyDiQA}), NER (\textsc{WikiANN}), and sentiment analysis (\textsc{IndicSentiment}, \textsc{AmazonReview})}.
{Each circle indicates performance given different $k$-shot demonstrations. 
There's a significant performance gap caused by the choice of demonstrations, which is often larger in ICL methods. }
} \label{fig:ablations}
\end{figure*}
\begin{figure*}[t!]
\centering
\includegraphics[width=16cm]{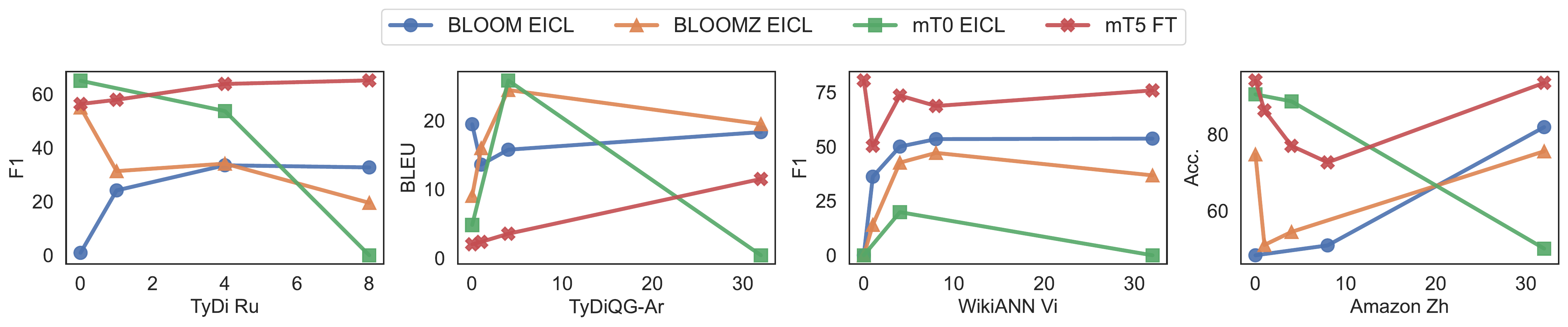}\caption{
\textbf{Demonstration scaling experiments on \textsc{TyDiQA} (Russian), \textsc{TyDiQA-QG} (Arabic), WikiANN (Vietnamese), and \textsc{Amazon Review} (Chinese) for four different models}. {$x$-axis indicate the number of $k$ demonstrations. While fine-tuning and ICL with pretrained LMs often benefit from fine-tuning, few-shot ICL with instruction-tuned models can result in performance deterioration. }
} \label{fig:demo_scaling}
\end{figure*}

%% file: sections/analysis.tex
\subsection{Analysis}
\label{sec:analysis}
\paragraph{Performance variance among different $k$ shots.}
Figure~\ref{fig:ablations} shows model performance across the three different $k$-shots and reveals a significant performance disparity in many of the tasks and languages. 
We observe the significant variance in fine-tuning-based transfer across different $k$-shot samples, confirming \cite{zhao-etal-2021-closer}.
Importantly, we show that in-context learning is even {\it more sensitive} to different demonstrations than few-shot fine-tuning. 
For instance, for \textsc{Amazon Review}, the standard deviation for BLOOM \textsc{E-CIL} and mT5 \textsc{Eng.+Tgt.} fine-tuning is 2.2 and 0.2, respectively.
We also analyze whether a demonstration set $k$ that achieves the best performance with a model also leads to the optimal performance for another model. 
Specifically, we compare the best $k$-shots for each task and language for BLOOM and BLOOMZ English ICL. We found that in 49.7\% of the cases, their optimal $k$-shot demonstrations differ. 
These results emphasize the difficulty of comparing model performance in the absence of standardized $k$-shot samples. 
On the bright side, these results provide insights into potential approaches for identifying optimal demonstrations that can enhance few-shot ICL performance.

\paragraph{{The effects of varying number of $k$. }}
{
{
Figure~\ref{fig:demo_scaling} demonstrates the impact of increasing the number of few-shot samples for in-context learning and fine-tuning, on four tasks: \textsc{TyDiQA}, \textsc{TyDiQA-QG}, \textsc{WikiANN}, and \textsc{Amazon Review}.  
Full results on the four tasks in a subset of the languages are available in Appendix~\ref{sec:more_varying_demo}. 
}
Specifically, we vary the number of few-shot demonstrations, including 1, 4, and 8 (for the tasks with more than 8 shots), and assess the performance of BLOOM~\textsc{English ICL}, BLOOMZ ~\textsc{English ICL}, mT0 \textsc{English ICL} and mT5 \textsc{Eng.+Tgt. FT}. 

Increasing the number of few-shot examples has a notable positive impact on fine-tuning (mT5 fine-tuning) across different tasks.  
Similarly, non-instruction-tuned BLOOM also benefits from the inclusion of few-shot samples on most of the tasks. 
However, for instruction-tuned models (mT0 and BLOOMZ), we observe a significant decline in performance when additional demonstrations are added, which aligns with the findings in Table~\ref{tab:main_results_light}. 
Specifically, on mT0, we observe that the zero-shot performance surpasses the few-shot performance on \textsc{TyDiQA} and \textsc{Amazon Review}. Surprisingly, even on previously unseen tasks such as \textsc{TyDiQA-QG} and \textsc{WikiANN}, the addition of more than four demonstrations leads to a significant decline in performance. 

It is worth noting that mT0 and BLOOMZ were exclusively trained with instructions and did not utilize demonstrations during training~\cite{muennighoff2022crosslingual}.
We hypothesize that this training approach may cause the models to overfit the zero-shot instruction-based in-context learning scenario, thereby hindering their ability to effectively learn in-context information through few-shot demonstrations. 
{\citet{wei2021finetuned} also find that while few-shot demonstrations mitigate high variance of the zero-shot inference with instructions only, the optimal zero-shot performance with the best template often outperforms the best few-shot performance.  
}

 
\paragraph{Effects of model scaling on few-shot in-context cross-lingual transfer.}
Figure~\ref{fig:scale_experiments} shows BLOOM-560 million, 1 billion, and 7 billion performance on a subset of the tasks. The transfer method is \textsc{English ICL}. 
As the model scales, the overall performance on few-shot in-context learning significantly improves, as found in English~\cite{brown2020language}, indicating that models' cross-lingual few-shot transfer performance via in-context learning may improve as the model size increases.
{
These findings are consistent with the results reported by \citet{xglm} on a set of classification tasks. 
}

\begin{figure}[t!]
  \centering
  \includegraphics[width=0.45\textwidth]{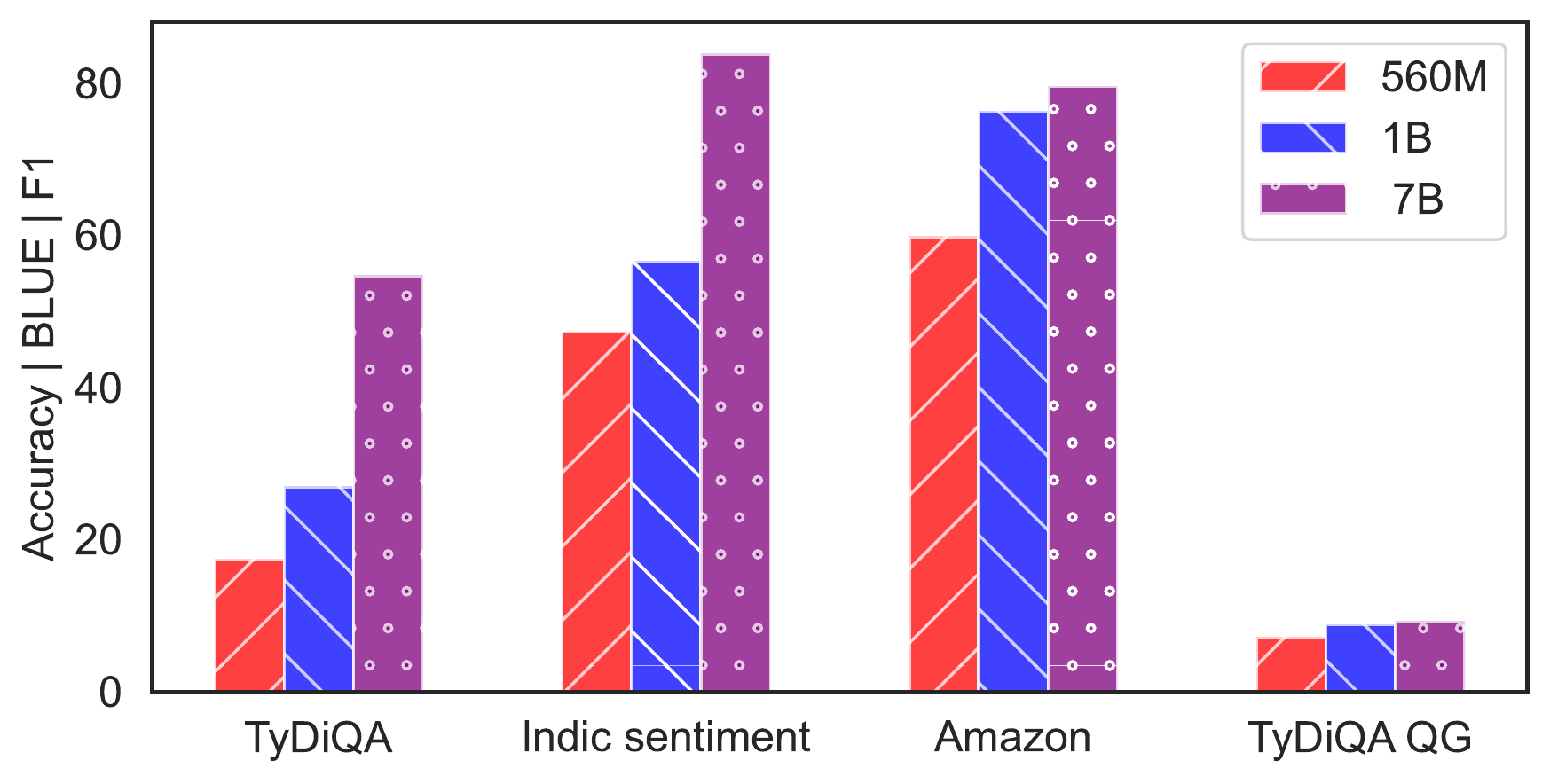}\label{fig:scaling experiments}
  \caption{\textbf{Model scaling experimental results.} We conduct experiments on four sub-tasks and use three BLOOM models, BLOOM-560M, 1B, and 7B. }
  \label{fig:scale_experiments}
\end{figure}

%% file: sections/discussion.tex
In this work, we introduce \data, a few-shot cross-lingual transfer benchmark that encompasses a diverse range of discriminative and generative tasks across a variety of typologically distinct languages.
Through our comprehensive evaluation, involving six different transfer methods and various LLMs, we offer valuable insights into the strengths and limitations of these transfer methods and LLMs. 
Our analysis reveals that while LLMs utilizing in-context learning excel in generation tasks, they are often surpassed by smaller fine-tuned models specifically trained for target tasks.
Furthermore, our findings highlight significant performance variations dependent on different transfer setups (e.g., demonstrations). 

Moving forward, our findings suggest the following exciting opportunities for future research in the field of few-shot learning transfer across diverse languages: 

\paragraph{Improve multilingual instruction tuning.}
Although instruction tuning can be beneficial for both zero-shot transfer, certain models, such as mT0, may become overly specialized for zero-shot instruction-tuning scenarios, leading to lower average few-shot performance than the optimal zero-shot performance. 
Although these models demonstrate impressive zero-shot performance, even on tasks they haven't encountered before (such as XCOPA), they face challenges when it comes to tasks that involve generating outputs in less commonly used formats (like structured predictions).
We believe that developing multilingual instruction-following models capable of effectively utilizing both instructions and demonstrations is crucial.
Recent studies demonstrate that incorporating both instructions and demonstrations during instruction-tuning on English data can enhance the model's performance~\cite{chung2022scaling}, allowing it to learn within context~\cite{min2021metaicl}. 
This type of training may potentially mitigate the issue of overfitting to specific formats. 
Hence, it is necessary to explore various instruction-tuning setups to further improve few-shot in-context learning, with a focus on {\it cross-lingual transfer}. 

Additionally, while high-quality human-translated instructions are effective, numerous instruction repositories are still dominated by English instructions. Therefore, community efforts to increase the availability of multilingual instructions may assist in the development of more generalizable multilingual large-language models.

\paragraph{Overcome data scarcity using LLMs.}
Our research reveals that smaller task-specific fine-tuned models, with intermediate training in English, can still outperform ChatGPT on discriminative tasks that require strict output formats. 
Conversely, ChatGPT outperforms fine-tuned models on tasks that necessitate more open-ended generations, such as question generation.
In recent studies, InstructGPT~\cite{ouyang2022training} has exhibited the ability to generate high-quality generations in English, even outperforming humans on some tasks \cite{goyal2022news}. 
This impressive capacity for flexible generations has prompted active investigations into generating training instances from such LLMs, which have predominantly focused on English \cite{wang2022self,honovich2022unnatural}. 
Some preliminary attempts have been made to explore task-specific data generation in certain target tasks, such as question answering \cite{agrawal2022qameleon}. 
However, there remains limited exploration on how to generate diverse task instructions and outputs for a variety of typologically diverse languages.
We believe that using LLMs to generate data offers a promising solution to obtaining more annotated data for under-represented languages.

\paragraph{Understand transfer dynamics in cross-lingual in-context learning.}
The impact of various instructions and demonstrations has been extensively examined in the context of English in-context learning, highlighting critical concerns such as sensitivity to prompt order \cite{lu-etal-2022-fantastically} and/or motivating methods for identifying optimal demonstrations \cite{su2022selective}.
This research has found that demonstrations or instructions that are optimal for one model may not necessarily result in the best performance for another model.
We anticipate that our benchmark will inspire and assist in further research into the relationship between language and instruction/demonstration for cross-lingual in-context learning. 

\paragraph{{Fairness beyond languages: underrepresented variants, dialects, and cross-cultural NLP.}}
{
Many of the diverse world languages are often excluded in widely used cross-lingual evaluation benchmarks, where recent papers show strong cross-lingual transfer capabilities. 
However, through our comprehensive analysis, we have discovered that even the most advanced LLMs currently available still face difficulties when dealing with less-represented languages.  
The most competitive instruction-tuned models, ChatGPT or mT0, show significant performance declines when it comes to indigenous languages, reaching a level akin to a random baseline.} 

We advocate for conducting more studies on diverse local languages, including under-represented languages and their dialects, as emphasized in previous works such as \citet{aji-etal-2022-one,kakwani-etal-2020-indicnlpsuite}. 
We note that datasets in such languages are often translated from English~\cite{yu2022counting}, which may introduce translation biases~\cite{artetxe-etal-2020-call} and fail to capture the linguistic nuances and interests of native speakers~\cite{clark-etal-2020-tydi,asai-etal-2021-xor}. 
To address these challenges, it is important that further work be done to develop cross-cultural Natural Language Processing ~\cite{hershcovich-etal-2022-challenges}. 

\paragraph{Expand evaluations to complex tasks.}
Most recent research on multilingual in-context learning predominantly focuses on discriminative tasks \cite{muennighoff2022crosslingual,ahuja2023mega} or translation tasks~\cite{xglm}. 
Further exploration can expand these evaluations to more diverse and complex tasks, such as MTOP \cite{li-etal-2021-mtop} or MGMS8K \cite{Shi2023language}, or knowledge-intensive tasks~\cite{asai-etal-2021-xor} as new multilingual benchmarks are developed.

%% file: sections/appendix.tex
\section{Benchmark Details  }
\subsection{\data Constructions}
This section will provide further details of the \data construction process. 

\paragraph{List of the languages.}
We show the list of the 55 languages included in \data in Table~\ref{tab:full_lang_list}. \data~covers 25 different language families, and also shows geographical diversities.

\paragraph{Examples.}
Table~\ref{tab:task_example} shows the input and output examples in \data. We reformulate all o the tasks with diverse formats into the same text-to-text format. 

\paragraph{Instructions.}
The full list of the instructions written in English is available in Table~\ref{tab:instructions_list}. 

\begin{table}[t!]
\centering
\footnotesize
\begin{tabular}{lll}
\toprule
                    Language name &                 Language family &  code \\
\midrule
              Amharic &      Afro-Asiatic & amh \\
               Arabic &      Afro-Asiatic &  ar \\
             Assamese &     Indo-European &  as \\
               Aymara & aymaran languages & aym \\
           Belarusian &     Indo-European &  be \\
              Bengali &     Indo-European &  bn \\
                 Boro &      Sino-Tibetan & brx \\
            Bulgarian &     Indo-European &  bg \\
               Bribri &          Chibchan & bzd \\
              Chinese &      Sino-Tibetan &  zh \\
            Asháninka &          Arawakan & cni \\
             Estonian &            Uralic &  et \\
              Finnish &            Uralic &  fi \\
               French &     Indo-European &  fr \\
               German &     Indo-European &  de \\
              Guarani &            Tupian &  gn \\
             Gujarati &     Indo-European &  gu \\
              Haitian &     French Creole &  ht \\
                Hausa &       Niger–Congo & hau \\
             Wixarika &       Uto-Aztecan & hch \\
                Hindi &     Indo-European &  hi \\
                 Igbo &       Niger–Congo & ibo \\
           Indonesian &      Austronesian &  id \\
              Italian &     Indo-European &  it \\
             Japanese &           Japonic &  ja \\
              Kannada &         Dravidian &  kn \\
          Kinyarwanda &       Niger–Congo & kin \\
               Korean &          Koreanic &  ko \\
                  Luo &      Nilo Saharan & luo \\
             Maithili &     Indo-European & mai \\
            Malayalam &         Dravidian &  ml \\
              Marathi &     Indo-European &  mr \\
 Modern Greek &     Indo-European &  el \\
              Nahuatl &       Uto-Aztecan & nah \\
Oriya (macrolanguage) &     Indo-European &  or \\
                Otomí &      Oto-Manguean & oto \\
              Panjabi &     Indo-European &  pa \\
      NigerianPidgin &    English Creole & pcm \\
              Persian &     Indo-European &  fa \\
           Portuguese &     Indo-European &  pt \\
              Quechua &            others &  qu \\
              Russian &     Indo-European &  ru \\
       Shipibo-Konibo &            Panoan & shp \\
              Spanish &     Indo-European &  es \\
               Swahil &       Niger–Congo &  sw \\
                Tamil &         Dravidian &  ta \\
             Rarámuri &       Uto-Aztecan & tar \\
               Telugu &         Dravidian &  te \\
                 Thai &           Kra–Dai &  th \\
              Turkish &            Turkic &  tr \\
                 Urdu &     Indo-European &  ur \\
           Vietnamese &     Austroasiatic &  vi \\
                Wolof &      Niger–Congo" & wol \\
               Yorùbá &       Niger–Congo & yor \\
\bottomrule
\end{tabular}
    \caption{List of all languages in \data. 
    }
    \label{tab:full_lang_list}
\end{table} 
    
\begin{table*}[t!]
\small
    \centering    
    \renewcommand{\arraystretch}{1.5}
    \begin{tabular}{p{2cm}| p{2cm} | p{8cm}| p{3cm} }
\toprule
Task & Dataset &  Input & Output \\\midrule
\rowcolor{red!10}
\textsc{NLI} & \textsc{Americas NLI} & premise: Ramonar mayamp jawsañaxanawakunalaykutix mä jiskt'aw utjitana …walikiwa…tukt'ayayita.. mä jisk't'aw utjitana kuntix lurkan ukata. \texttt{[SEP]} hypothesis: Janiw jayraskayat Ramonar jawsañxa. (aym) & \texttt{contradiction} \\ 
 \rowcolor{orange!20}
\textsc{Paraphrase} & \textsc{PAWS-X} & sentence 1: Ses parents sont Angelina Miers, une artiste de premier plan, et Don Luis Toranzos, d’Argentine. \texttt{[SEP]} sentence 2: Ses parents sont Angelina Miers, elle-même un artiste de premier plan, et Don Luis Toranzos d'Argentine. (fr) & \texttt{duplicate} \\
\rowcolor{yellow!20}
\textsc{Sentiment}  & \textsc{Amazon} &  review title: \zhtext{质量很好，空间容量大，可以装很多东西 review body: 箱子很轻盈，柔韧性不错，不易变形。外观优雅美观，出行很有范，呵呵。好评！} (zh) & \texttt{positive} \\ 
\rowcolor{green!10}
\textsc{Commonsense} & \textsc{XCOPA} & Õpetaja andis õpilastele kodutöö. \texttt{(A)} Õpilased saatsid kirju. \texttt{(B)} Õpilased ägisesid. (et) & \texttt{(B)} \\
\rowcolor{green!10}
\textsc{Commonsense} & \textsc{XWINOGRAD} & \jatext{フリースは綿より感触がよい。\_のほうがずっと柔らかいからいだ。} \texttt{(A)} \jatext{フリース} \texttt{(B)} \jatext{綿}& \texttt{(A)} \\
  \rowcolor{blue!10}
\textsc{QA} & \textsc{TyDiQA} & question: Mikä oli Stanley Kubrickin ensimmäinen elokuva? context: Lyhytelokuvien jälkeen Kubrick teki ensimmäisen pitkän elokuvansa Fear and Desire vuonna 1953 rahoittaen sen kokonaan itse ja sukulaistensa avustuksella, mikä oli tuolloin hyvin epätavallista. Kubrickin esikoiselokuva oli kuitenkin floppi, ja ohjaaja osti kaikki esityskopiot itselleen, jotta elokuvaa ei näytettäisi. Elokuva merkitsi myös hänen ensimmäisen avioliittonsa loppua, koska Kubrick tapasi kuvauksien aikana Ruth Sabotkan, jonka kanssa hän muutti yhteen avioeronsa jälkeen. Kubrick ja Sabotka menivät naimisiin vuonna 1955, ja he saivat yhdessä yhden lapsen, Katharinan (syntynyt 1953).	(fi) & Fear and Desire \\ 
  \rowcolor{purple!10}
\textsc{NER}  & \textsc{MasakhanER}  &Issachar alikuwa ametokea India akielekea Israel ambapo aliwekwa chini ya ulinzi na hakutakiwa kutoka nje ya uwanja wa ndege wa Russia . & India \texttt{<organization>} Israel \texttt{<organization>} Russia \texttt{<organization>} \\ 
   \rowcolor{gray!10}
\textsc{QG} & \textsc{TyDIQA-QG} & premise: \kotext{롯데는 이번 상반기 채용과 관련해 구직자들에게 실질적인 도움이 될 수 있도록 다양한 방법으로 정보제공 활동을 강화할 계획이다. [SEP] hypothesis: 롯데는 어떠한 정보도 제공하지 않을 계획이다.} & \texttt{contradiction} \\
\bottomrule
 \end{tabular}
    \caption{The input and output examples in \data. We show one example from one dataset per task. Due to the long input length, we do not include a summarization example in this table. 
    }
    \label{tab:task_example}
\end{table*}

\begin{table*}[t!]
\small
    \renewcommand{\arraystretch}{1.5}
    \centering
    \begin{tabular}{p{2cm}| p{13cm} }
\toprule
Dataset & Instructions \\\midrule
\rowcolor{red!10}
\textsc{NLI} & Take the premise sentence as truth. Then the hypothesis is true (entailment), false (contradiction) or inconclusive (neutral)? \\
 \rowcolor{orange!20}
\textsc{PAWS-X}  & In this task you are given a sentence pair that has high lexical overlap. If the sentences have the same meaning and are just paraphrases of each other label them as duplicate, if not label them as not\_duplicate \\
\rowcolor{yellow!20}
\textsc{Sentiment} & In this task, you're given a review from Amazon. Your task is to generate a rating for the product. The rating is  extremely negative, negative, neutral, positive, and really positive. \\ 
\rowcolor{green!10}
\textsc{XCOPA} &  In this task you are given a premise and two alternatives (A) and (B). You must choose the alternative that is more plausibly the cause or effect of the situation described by the premise. \\
\rowcolor{green!10}
\textsc{XWINOGRAD} &  Replace the \_ in the above sentence with the correct option  \\ 
  \rowcolor{blue!10}
QA & Read the given passage and answer a question about the information present in the passage. \\ 
  \rowcolor{purple!10}
\textsc{NER} & Given the following sentence, indicate the name entities (i.e., the real-world objects such as a person, location, organization, etc. that can be denoted with a proper name) such as ``New York Times''. For each word of a named-entity, indicate their type ``location'' or ``organization'' or ``person''.  \\
   \rowcolor{gray!10}
\textsc{Summarization} &  In this task, you are given an article. Your task is to summarize the article in a sentence. \\
   \rowcolor{gray!10}
\textsc{QG} &  This task is about reading the given passage and constructing a question about the information present in the passage. \\
\bottomrule
 \end{tabular}
    \caption{The list of English instructions for each task in \data. 
    }
    \label{tab:instructions_list}
\end{table*}

\paragraph{List of the datasets with languages included.}
Table~\ref{tab:buffet_full_lang_list} shows the full list of the datasets with language names included in \data. 

\subsection{\data-Light}
\label{sec:buffet_light}
\paragraph{Task and language decisions.}
The goal of building the \data-Light subset is to enable quick multilingual evaluation without losing the language and task diversity in the original \data. To this end, we filter \data so that we evaluate between 3 and 7 languages per task, and each language is included in at most three tasks.\footnote{In addition to the high-resource languages per task, we also include low-resource languages when available (i.e., for \textsc{NLI}) to not unfairly inflate \data-Light scores.} This design choice allows us to consider 31 diverse languages across all tasks in BUFFET while reducing the number of evaluation settings by 66\%.

\paragraph{Final list of \data-light.}
The full list of tasks and languages in \data are in Table~\ref{tab:buffet_light_lang_list}.  

\begin{table*}[t!]
\small
    \renewcommand{\arraystretch}{1.}
    \centering
    \begin{tabular}{l| l |p{8cm} }
\toprule
Task & Dataset & Languages \\\midrule
\rowcolor{red!10}
\textsc{NLI} &  \textsc{Americas NLI} & aym, bzd, cni, gn, hch, nah, too, quy, shp, tar \\
\rowcolor{red!10}
 & \textsc{KLUE NLI} & ko \\
 \rowcolor{red!10}
 & \textsc{OCNLI} & zh \\
 \rowcolor{red!10}
  & \textsc{Parsi NLU entailment} & fa \\
  \rowcolor{red!10}
    & XNLI   & ar, bg, de, el, en, es, fr, hi, ru, sw, th, tr, ur, vi, zh \\
     \rowcolor{orange!20}
\textsc{Paraphrase Detection}  & \textsc{PAWS} & (en,) de, es, fr, ja, ko, zh \\
\rowcolor{yellow!20}
\textsc{Sentiment} & \textsc{Amazon Review} & (en), de, es, fr, ja, zh \\ 
\rowcolor{yellow!20}
\textsc{Analysis} & \textsc{Indic Sentiment} & as, bn, brx, gu, hi, kn, mai, ml, mr, or, pa, ta, te, ur \\
\rowcolor{green!10}
\textsc{Commonsense} & \textsc{XCOPA} & et, ht, it, id, qu, sw, zh, ta, th, tr, vi  \\
\rowcolor{green!10}
\textsc{Commonsense} & \textsc{XWINOGRAD} & (en,) ja, pt, ru, zh   \\ 
  \rowcolor{blue!10}
\textsc{QA} &  \textsc{TyDiQA} & (en,) ar, be, fi, id, sw, ko, ru, te \\ 
 \rowcolor{purple!10}
\textsc{NER} & \textsc{WikiAnn} & ( en,) ta, fr, it, ja, vi, qu, be, gu, et, th, or, kn, fi, gn, ru, el, ur, es, hi, te, as, sw, pa, bg, ml, tr, fa, id, ko, mr, de, ar, bn, zh \\
 \rowcolor{purple!10}
 & \textsc{MasakhaNER} &  amh, hau, ibo, kin, luo, pcm, swa, wol, yor \\ 
   \rowcolor{gray!10}
\textsc{Summarization} & \textsc{XLSum} & (english, ) ta, vi, id, tr, ja, th, bn, ar, en, es, fa, zh, sw \\
   \rowcolor{gray!10}
\textsc{QG} & \textsc{TyDiQA-QG}  & (en,) ar, be, fi, id, sw, ko, ru, te \\
\bottomrule
 \end{tabular}
    \caption{The list of datasets with language lists in \data. 
    }
    \label{tab:buffet_full_lang_list}
\end{table*}

\begin{table*}[t!]
\small
    \renewcommand{\arraystretch}{1.}
    \centering
    \begin{tabular}{l| l |p{8cm} }
\toprule
Task & Dataset & Languages \\\midrule
\rowcolor{red!10} NLI & \textsc{Americas NLI} & aym, cni, hch \\
\rowcolor{red!10} & \textsc{KLUE NLI} & ko \\
\rowcolor{red!10} & \textsc{Parsi NLU Entailment} & fa \\
\rowcolor{red!10} & \textsc{XNLI} & bg, el, hi, sw, ur \\
\rowcolor{orange!20} Paraphrase Detection  & \textsc{PAWS-X} & de, es, ja, ko, zh \\
\rowcolor{yellow!20} Sentiment & \textsc{Amazon Review} & de, fr, ja, zh \\ 
\rowcolor{yellow!20} Analysis & \textsc{Indic Sentiment} & bn, ta, ur \\
\rowcolor{green!10} Commonsense & \textsc{XCOPA} & et, it, ta, th, tr  \\
\rowcolor{green!10} & \textsc{XWINOGRAD} & pt, ru \\ 
\rowcolor{blue!10} QA & \textsc{TyDiQA} & be, id, sw \\ 
\rowcolor{purple!10} NER & \textsc{WikiAnn} & be, bg, el, et, fi, it \\ 
\rowcolor{purple!10} & \textsc{MasakhaNER} & yor \\ 
 \rowcolor{gray!10} Summarization & \textsc{XLSum} & bn, fa, es, id, tr, vi \\ 
 \rowcolor{gray!10} QG & \textsc{TyDiQA-QG}  & ar, fi, ko, ru, te \\
\bottomrule
 \end{tabular}
    \caption{The subset of datasets and languages included in \data-Light.
    }
    \label{tab:buffet_light_lang_list}
\end{table*}

\section{More Experimental Details}
\paragraph{Fine-tuning. }

For \textsc{English FT}, we limit the number of English training samples to 100,000 and fine-tune mt5-base~\cite{xue2020mt5} for 3 epochs. 
For the \textsc{English FT} baseline, we transfer this model directly to new languages, while for 
\textsc{Eng.+Tgt. FT}, we initialize the model checkpoint with the trained model weight and further fine-tune a model on few-shot samples for 300 epochs. 

\paragraph{In-context learning. }
We set the maximum token length to 15 except for \textsc{XLSum} and \textsc{TyDiQA-QG}. 
For \textsc{XLSum}, we set the maximum token length to 100, and for \textsc{TyDiQA-QG}, we set the maximum token length to 50. 
We use greedy decoding throughout the experiments. 
For BLOOM-based model evaluations, we use a single RTX-100 GPU with 24 GB GPU memory. 
We use int8bit quantization to avoid GPU out-of-memory errors. To evaluate mT5 and mT0, we use TPU v3-8.

\begin{table*}[t!]
\small
    \centering
    \renewcommand{\arraystretch}{1.3}
    \begin{tabular}{ l|r|ccccccccccc}
\toprule
  &  Output &  \multicolumn{3}{c}{Classification} &  \multicolumn{2}{c}{Multiple Choice} & Span & Str. &  \multicolumn{2}{c}{Generation} &  \multicolumn{2}{c}{Avg.}  \\

&Tasks & NLI& Sent. & Para. &XCPA&XWGD &  QA  & NER & QG& Summ. & class & gen \\\midrule
Random &  & 33.3 & 50.0 & 50.0 & 50.0&  50.0 & -- & -- & -- & -- & -- & \\\hline
\textsc{Tgt. FT} & {mT5} &34.6 &                67.2 &                  47.2 &   46.7 &       50.0 &                 8.3 &                    30.8 &                  3.4 &            2.8 &              40.2 &             3.1   \\
\textsc{Eng. FT} &  {mT5} & 46.0 &                89.7 &                  78.6 &    0.0 &        0.0 &                62.9 &                    30.8 &                  4.2 &            4.0 &             48.2 &             4.1 \\ 
\textsc{Eng.+Tgt.} & mT5 & {\bf 48.8} &                 {\bf 90.4} &                  {\bf 77.9} &   49.9 &       49.0 &               {\bf 66.7} &                    {\bf 43.5} &                 12.2 &          {\bf  8.4} &              {\bf 58.8} &           {\bf  10.0} \\
\midrule
\textsc{Eng. ICL}& BLOOM&     33.6 &                85.3 &                  42.4 &   50.0 &       50.8 &                39.2 &                    25.0 &                 11.6 &            2.4 &              44.0 &             7.0 \\
& mT5&  34.5 &                50.0 &                  43.2 &   50.0 &       49.2 &                 0.3 &                     1.6 &                  0.0 &            0.3 &              32.1 &             0.1 \\
  \rowcolor{blue!10}
& BLOOMZ&  33.0 &                87.2* &                  49.5* &   50.5 &       52.1 &                44.5* &                    20.0 &                 13.9 &            9.0* &              44.3 &            11.4 \\
  \rowcolor{blue!10}
& mT0 & 33.6 &                79.9* &                  61.1* &   57.1 &       59.6 &                69.0* &                     7.9 &                 15.3 &            1.5* &              45.6 &             8.4  \\
  \rowcolor{blue!10}
 & ChatGPT$\dagger$ &   {\bf \underline{54.5}} &                91.1 &                  68.6&  {\bf \underline{ 76.7 }} &       73.3 &                68.1 &                   {\bf \underline{ 45.4 }}&                 {\bf \underline{21.2}} &            5.4 &              {\bf \underline{64.6}} &            13.3 \\ 
\hdashline
\textsc{Tgt. ICL}& BLOOM &     31.7 &                85.3 &                  45.9 &   50.1 &       51.7 &                 7.0 &                    25.2 &                {  12.8} &            4.7 &              41.2 &             8.7 \\
& mT5&   34.4 &                50.0 &                  43.1 &   50.0 &       47.3 &                 0.2 &                     0.2 &                  0.0 &            0.3 &              31.7 &             0.1 \\
  \rowcolor{blue!10}
& BLOOMZ &  32.1 &                64.7* &                  51.7* &   50.5 &       53.1 &                43.7* &                    19.1 &                 12.0 &           10.9* &              40.6 &            11.4 \\
  \rowcolor{blue!10}
& mT0& 38.1 &                70.6* &                  60.9* &   52.8 &       57.9 &                70.8* &                     8.5 &                 14.6 &            1.8* &              45.7 &             8.2   \\
  \rowcolor{blue!10}
& ChatGPT$\dagger$ &48.2 &               {\bf \underline{ 91.5}} &                  68.2 &   74.3 &       {\bf \underline{73.4}} &                68.0 &                    44.8 &                 21.1 &           {\bf \underline{11.4}} &              62.7 &           {\bf \underline{  16.3}} \\
\hdashline
\textsc{Z-EICL}& BLOOM&   32.3 &             35.8 &          42.3&         50.1  &  46.4  &      3.1  &  0.0  &                   {\bf 16.4} &                4.1 &           28.8&             10.0  \\
& mT5& 34.2&               50.0  &                  42.4 &   50.1 &       46.4 &                 2.0 &                     0.0 &                 0.1 &       1.3 &              32.5 &            0.7 \\
  \rowcolor{blue!10}
& BLOOMZ & 34.0 &                51.6* &                  58.0* &   50.1 &       50.9 &                65.2* &                     7.6 &                 10.2 &            2.9* &              39.3 &             6.6  \\
  \rowcolor{blue!10}
& mT0&   49.1 &                90.2* &                  91.2* &   64.1 &       64.5 &                75.2* &                     0.0 &                 10.3 &            8.5* &              56.0 &             9.4 \\
\bottomrule
 \end{tabular}
    \caption{\textbf{Overall experiment results on \data}.  
    {\hlc[blue!10]{The blue-colored rows} are instruction-tuned models, and we added $^*$ symbols next to the scores for the tasks on which the models have been trained. 
    ``Random'' shows random baseline performance.
    } 
    {\bf Bold} fonts indicate the best results for each task, among the models that are not directly trained on the task.  
    When ChatGPT achieves the best results, we also note the second-best number from the models that are not trained on the task, acknowledging the possibility that ChatGPT may have encountered a similar task during training.
    }
    \label{tab:main_results}
\end{table*}

\section{Detailed \data Results}
This section includes the full list of the experimental results. 
Overall results on the full \data are available in Table~\ref{tab:main_results}, and 
Figure~\ref{tab:main_results_light} summarize overall performance across the eight tasks, on the \data-Light subset. 
The overall trends on \data-light remain the same as the original \data. 
This indicates \data-Light is a reliable and more efficient alternative for holistic evaluations for few-shot cross-lingual transfer. 

Below, we present the performance breakdown for each dataset. 

\begin{figure}[t!]
\centering
\includegraphics[width=8cm]{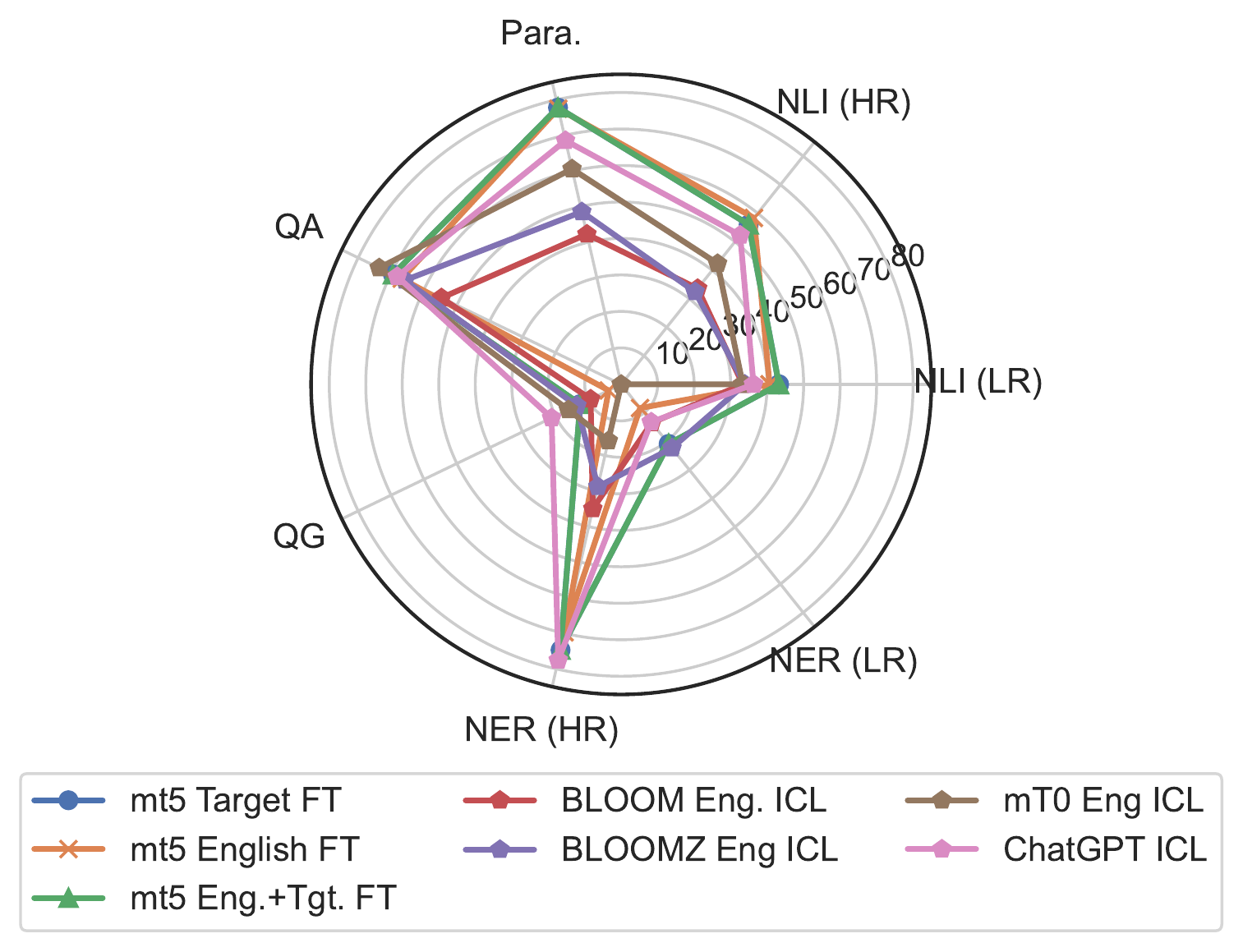}\caption{
Overall results on \data-Light.  
} 
\label{fig:overall}
\end{figure}

\subsection{NLI}
Table~\ref{tab:americas_nli} shows the full results on \textsc{AmericasNLI}. 
Table~\ref{tab:xnli} shows the full results on \textsc{XNLI}.
Table~\ref{tab:othernli_results} presents the full results on the other three entailment datasets annotated in each language, \textsc{KLUENLI}, \textsc{OCNLI}, and \textsc{ParsiNLUEntailment}. 

On \textsc{XNLI}, \textsc{English FT} (zero-shot transfer) shows strong performance and often outperforms \textsc{Eng.+Tgt. FT} (few-shot transfer). Among \textsc{ICL} baselines, mT0 \textsc{ZICL} shows the best macro performance on \textsc{XNLI}. 
However, on \textsc{AmericasNLI}, all methods struggle, while \textsc{Eng.+Tgt. FT} shows the best macro performance on \textsc{Americas NLI}. The performance gap between \textsc{English FT} and \textsc{Eng.+Tgt. FT} get significantly larger, with the largest gap in Aymara (5.5\%). 
Despite its strong performance on \textsc{XNLI}, mT0 \textsc{ZICL} struggles in  \textsc{Americas NLI} (33.7\% on average). 

While mT0 \textsc{ZICL} shows robust performance across \textsc{XNLI} languages, ChatGPT shows a large performance gap between higher-resource languages and low-resource languages (57\% in Greek v.s. 33\% Urdu). 

\begin{table*}[t!]
\footnotesize
\centering
\begin{tabular}{l|rrrrrrrrrrr}\toprule      Transfer + Model &  Macro &  aym &  bzd &  cni &   gn &  hch &  nah &  oto &  quy &  shp &  tar \\\midrule        Target FT &   35.9 & 36.0 & 35.5 & 35.5 & 35.7 & 32.7 & 37.5 & 35.2 & 35.4 & 37.6 & 37.8 \\       English FT &   42.6 & 40.7 & 44.9 & 43.3 & 46.8 & 38.0 & 42.5 & 41.6 & 46.1 & 43.2 & 39.2 \\English Target FT &   45.1 & 46.2 & 48.6 & 45.0 & 49.7 & 38.8 & 46.8 & 44.2 & 46.4 & 42.5 & 43.0 \\       EICL BLOOM &   33.7 & 33.4 & 34.6 & 33.2 & 34.1 & 33.3 & 33.5 & 33.4 & 34.3 & 34.0 & 33.6 \\         EICL mT5 &   33.3 & 33.3 & 32.8 & 33.3 & 33.3 & 33.2 & 33.2 & 33.2 & 33.3 & 33.3 & 33.3 \\      EICL BLOOMZ &   33.3 & 33.1 & 33.5 & 33.7 & 33.3 & 33.3 & 33.8 & 32.0 & 33.3 & 33.3 & 33.3 \\         EICL mT0 &   33.3 & 33.3 & 33.2 & 33.3 & 33.3 & 33.4 & 33.3 & 33.3 & 33.4 & 33.3 & 32.9 \\     EICL ChatGPT &   36.3 & 33.6 & --& 40.9 & --& 34.3 & --& --& --& --& --\\       TICL BLOOM &   33.7 & 33.5 & 34.6 & 33.2 & 33.6 & 33.3 & 33.5 & 33.3 & 34.3 & 34.0 & 33.6 \\         TICL mT5 &   33.3 & 33.3 & 32.8 & 33.3 & 33.6 & 33.2 & 33.2 & 33.3 & 33.3 & 33.3 & 33.3 \\      TICL BLOOMZ &   33.4 & 33.3 & 33.5 & 33.7 & 33.3 & 33.3 & 33.8 & 33.4 & 33.3 & 33.3 & 33.3 \\         TICL mT0 &   33.4 & 33.6 & 33.3 & 33.3 & 33.3 & 33.3 & 33.3 & 33.3 & 33.3 & 33.3 & 33.3 \\     TICL ChatGPT &   34.7 & 33.6 & --& 36.7 & --& 33.9 & --& --& --& --& --\\       ZICL BLOOM &   33.5 & 33.7 & 32.0 & 33.7 & 32.5 & 34.7 & 31.6 & 33.8 & 34.7 & 34.7 & 33.9 \\         ZICL mT5 &   34.0 & 36.3 & 34.4 & 32.9 & 32.8 & 36.4 & 33.6 & 33.7 & 32.9 & 33.3 & 34.1 \\      ZICL BLOOMZ &   34.3 & 36.3 & 33.5 & 33.7& 33.3 & 36.4 & 33.6 & 33.7 & 32.9 & 33.3 & 34.1 \\         ZICL mT0 &   33.7 & 33.5 & 33.5 & 33.3 & 33.7 & 33.3 & 34.1 & 33.2 & 35.3 & 33.1 & 33.5 \\\bottomrule\end{tabular}
     \caption{Model performance on \textsc{AmericasNLI}. We report the average of the three few-shot samples. 
    }
    \label{tab:americas_nli}
\end{table*}

\begin{table*}[t!]
    \centering
    \footnotesize
\begin{tabular}{l|rrrrrr}\toprule      Transfer + Model &  Macro &   ar &   bg &   de &   el &   es \\\midrule        Target FT &   36.4 & 35.8 & 37.8 & 37.3 & 37.4 & 37.0 \\       English FT &   59.4 & 59.2 & 62.9 & 61.5 & 61.4 & 63.7 \\English Target FT &   57.3 & 57.7 & 59.5 & 59.0 & 59.4 & 62.7 \\       EICL BLOOM &   33.7 & 34.0 & 33.9 & 33.4 & 33.3 &34.2 \\         EICL mT5 &   33.3 & 33.3 & 33.3 & 33.3 & 33.3 & 33.3 \\      EICL BLOOMZ &   33.1 & 34.1 & 33.6 & 33.7 & 27.9 & 34.2 \\         EICL mT0 &   36.3 & 37.8 & 36.3 & 35.3 & 33.4 & 33.7 \\     EICL ChatGPT &   50.3 & --& 60.7 & --& 54.0 & --\\       TICL BLOOM &   33.4 & 33.6 & 32.7 & 33.2 & 33.7 & 32.9 \\         TICL mT5 &   33.3 & 33.3 & 33.3 & 33.3 & 33.2 & 33.3 \\      TICL BLOOMZ &   33.4 & 33.3 & 33.7 & 33.3 & 34.4 & 33.3 \\         TICL mT0 &   40.4 & 38.8 & 51.2 & 41.8 & 47.8 & 43.1 \\     TICL ChatGPT &   50.5 & --& 52.4 & --& 56.9 & --\\       ZICL BLOOM &   33.6 & 33.7 & 34.1 & 34.3 & 33.7 & 33.7 \\         ZICL mT5 &   32.3 & 32.8 & 32.1 & 32.5 & 32.3 & 30.6 \\      ZICL BLOOMZ &   32.1 & --& --& --& --& --\\         ZICL mT0 &   56.2 & 56.1 & 58.4 & 58.7 & 57.5 & 58.0 \\\bottomrule\end{tabular}
\begin{tabular}{l|rrrrrrrrr}
\toprule      Transfer + Model &   fr &   hi &   ru &   sw &   th &   tr &   ur &   vi &   zh \\\midrule        Target FT & 37.4 & 35.7 & 36.0 & 35.1 & 36.7 & 36.8 & 34.2 & 36.3 & 35.5 \\       English FT & 62.1 & 58.0 & 59.8 & 55.5 & 57.4 & 58.4 & 54.0 & 57.1 & 60.4 \\English Target FT & 59.0 & 55.1 & 60.1 & 52.3 & 56.4 & 56.1 & 51.6 & 55.8 & 58.3 \\       EICL BLOOM & 36.2 & 33.4 & 33.6 & 33.4 & 33.3 & 33.3 & 33.3 & 33.3 & 33.4 \\         EICL mT5 & 33.4 & 33.3 & 33.3 & 33.3 & 33.3 & 33.3 & 33.3 & 33.3 & 33.3 \\      EICL BLOOMZ & 35.1 & 33.4 & 32.1 & 33.9 & 33.0 & 32.1 & 33.1 & 33.2 & 33.8 \\         EICL mT0 & 47.3 & 36.3 & 34.9 & 35.8 & 33.4 & 38.1 & 34.9 & 37.9 & 33.7 \\     EICL ChatGPT & --& 48.0 & --& 55.9 & --& --& 33.1 & --& --\\       TICL BLOOM & 33.3 & 33.3 & 33.2 & 34.3 & 34.8 & 33.8 & 33.6 & 32.5 & 33.0 \\         TICL mT5 & 33.3 & 33.2 & 33.3 & 33.3 & 33.5 & 33.3 & 33.3 & 33.3 & 33.3 \\      TICL BLOOMZ & 32.9 & 33.2 & 34.0 & 33.6 & 33.7 & 32.9 & 33.1 & 32.8 & 33.3 \\         TICL mT0 & 39.7 & 39.9 & 47.7 & 37.3 & 37.4 & 33.5 & 35.7 & 35.3 & 36.8 \\     TICL ChatGPT & --& 51.8 & --& 47.3 & --& --& 44.2 & --& --\\       ZICL BLOOM & 34.0 & 33.4 & 33.5 & 33.9 & 33.3 & 33.1 & 34.7 & 33.3 & 32.3 \\         ZICL mT5 & 29.6 & 33.3 & 32.3 & 32.7 & 33.1 & 34.7 & 32.8 & 32.4 & 31.1 \\      ZICL BLOOMZ & --& --& --& --& --& --& 32.8 & 32.4 & 31.1 \\         ZICL mT0 & 58.7 & 55.3 & 57.0 & 53.7 & 51.6 & 56.1 & 54.5 & 57.3 & 54.5 \\\bottomrule
\end{tabular}
     \caption{Model performance on \textsc{XNLI}. 
     We report the average of the three few-shot samples. 
    }
    \label{tab:xnli}
\end{table*}

\begin{table*}[t!]
    \centering
    \footnotesize
\begin{tabular}{l|rrr}
\toprule      Transfer + Model &  \textsc{KLUENLI} &  \textsc{ParsinluEntailment} &  \textsc{OCNLI} \\\midrule        Target FT &            34.0 &                       34.6 &   34.0 \\       English FT &            57.9 &                       51.1 &   32.5 \\English Target FT &            61.1 &                       50.5 &   38.6 \\       EICL BLOOM &            33.8 &                       28.9 &   38.9 \\         EICL mT5 &            33.3 &                       40.4 &   31.0 \\      EICL BLOOMZ &            31.9 &                       28.8 &   38.2 \\         EICL mT0 &            34.3 &                       30.0 &   36.7 \\     EICL ChatGPT &            64.8 &                       62.3 &   --\\       TICL BLOOM &            33.4 &                      28.8 &   38.2 \\         TICL mT5 &            33.3 &                       40.4 &   30.5 \\      TICL BLOOMZ &            33.8 &                       29.0 &   32.1 \\         TICL mT0 &            43.1 &                       37.4 &   38.6 \\     TICL ChatGPT &            56.5 &                       50.2 &   --\\       ZICL BLOOM &            33.8 &                       37.4 &   32.0 \\         ZICL mT5 &            32.4 &                       31.9 &   37.6 \\      ZICL BLOOMZ &            32.4 &                       31.9 &   37.6 \\         ZICL mT0 &            56.9 &                       55.2 &   50.6 \\\bottomrule\end{tabular}
     \caption{Model performance on \textsc{KLUENLI}, \textsc{OCNLI} and \textsc{ParsiNLUEntailment}. 
     We report the average of the three few-shot samples. 
    }
    \label{tab:othernli_results}
\end{table*}

\subsection{Paraphrase Detection}
The results on \textsc{PAWS-X} results are available in Table\ref{tab:pawsx_results}. 
 \textsc{Eng. FT} shows the best performance on this task among non-instruction-tuned models. 
 We hypothesize that as the languages included in \textsc{PAWS-X} are all relatively well-represented languages and the task is relatively simple,  \textsc{Eng. FT}, which is not trained in the target languages, can achieve high performance. 
mT0 \textsc{ZICL} shows quite high performance, likely because the model is trained on \textsc{PAWS-X}~\cite{muennighoff2022crosslingual}. 

\begin{table*}[t!]
    \centering
    \footnotesize
\begin{tabular}{l|rrrrrrr}\toprule      Transfer + Model &  Macro &   de &   es &   fr &   ja &   ko &   zh \\\midrule        Target FT &   47.2 & 47.5 & 48.8 & 47.1 & 48.1 & 44.2 & 47.3 \\       English FT &   78.6 & 79.9 & 83.5 & 84.0 & 74.5 & 74.3 & 75.5 \\English Target FT &   77.9 & 79.9 & 82.6 & 81.0 & 73.1 & 73.9 & 77.0 \\       EICL BLOOM &   42.4 & 41.5 & 42.3 & 43.0 & 42.7 & 42.0 & 42.8 \\         EICL mT5 &   43.2 & 41.5 & 42.4 & 47.7 & 42.7 & 42.0 & 42.6 \\      EICL BLOOMZ &   49.5 & 58.9 & 58.9 & 57.7 & 34.5 & 29.5 & 57.8 \\         EICL mT0 &   61.1 & 78.7 & 57.6 & 57.8 & 57.3 & 58.0 & 57.4 \\     EICL ChatGPT &   68.6 & 73.5 & 72.0 & --& 67.4 & 60.1 & 69.8 \\       TICL BLOOM & 45.9 & 49.3 & 42.3 & 42.4 & 42.9 & 54.9 & 43.0 \\         TICL mT5 &   43.1 & 41.5 & 46.4 & 43.0 & 42.7 & 42.0 & 42.6 \\      TICL BLOOMZ &   51.7 & 47.4 & 56.4 & 51.3 & 48.8 & 55.6 & 50.4 \\         TICL mT0 &   60.9 & 67.9 & 68.1 & 57.0 & 57.3 & 58.0 & 57.4 \\     TICL ChatGPT &   68.5 & 71.9 & 71.5 & --& 67.0 & 62.8 & 69.1 \\       ZICL BLOOM &   42.4 & 41.6 & 42.4 & 42.9 & 43.0 & 42.0 & 42.7 \\         ZICL mT5 &   58.0 & 58.0 & 57.8 & 58.6 & 57.7 & 58.1 & 57.5 \\      ZICL BLOOMZ &   58.0 & 58.0 & 57.8 & 58.6 & 57.7 & 58.1 & 57.5 \\         ZICL mT0 &   91.2 & 91.5 & 95.5 & 94.3 & 87.5 & 87.9 & 90.8 \\\bottomrule\end{tabular}
     \caption{Model performance on \textsc{PAWSX}. 
     We report the average of the three few-shot samples. 
    }
    \label{tab:pawsx_results}
\end{table*}

\subsection{Sentiment Analysis}
The experimental results on \textsc{Amazon Review Multi} and \textsc{Indic Sentiment} are available in Tables~\ref{tab:amazon_review_result} and \ref{tab:indic_results}. 
On both datasets, all models yield high accuracy across languages, except for mT5 \textsc{ZEICL}. 

\begin{table*}[t!]
    \centering
    \footnotesize
\begin{tabular}{l|rrrrrr}\toprule      Transfer + Model &  Macro &   de &   zh &   es &   fr &   ja \\\midrule        Target FT &   76.3 & 72.9 & 77.1 & 76.1 & 82.3 & 73.1 \\       English FT &   91.9 & 94.2 & 84.5 & 93.8 & 95.1 & 91.8 \\English Target FT &   92.4 & 93.6 & 87.6 & 93.4 & 94.9 & 92.3 \\       EICL BLOOM &   83.4 & 82.0 & 84.9 & 92.8 & 88.0 & 69.2 \\         EICL mT5 &   50.2 & 49.4 & 50.6 & 50.9 & 50.6 & 49.8 \\      EICL BLOOMZ &   81.5 & 75.7 & 80.2 & 93.8 & 93.5 & 64.3 \\         EICL mT0 &   79.8 & 88.7 & 70.6 & 81.8 & 89.6 & 68.5 \\     EICL ChatGPT &   85.8 & 94.3 & 87.5 & --& 96.1 & 65.0 \\       TICL BLOOM &   84.2 & 87.3 & 85.7 & 92.8 & 84.2 & 70.9 \\         TICL mT5 &   50.2 & 49.4 & 50.6 & 50.9 & 50.6 & 49.8 \\      TICL BLOOMZ &   64.9 & 57.1 & 71.2 & 79.2 & 61.5 & 55.5 \\         TICL mT0 &   72.2 & 88.9 & 51.3 & 58.9 & 85.1 & 76.8 \\     TICL ChatGPT &   89.7 & 94.4 & 85.5 & --& 95.6 & 83.2 \\       ZICL BLOOM &   50.3 & 49.4 & 50.6 & 50.9 & 50.7 & 49.8 \\         ZICL mT5 &   45.1 & 48.5 & 49.6 & 39.9 & 37.0 & 50.4 \\      ZICL BLOOMZ &   15.6 & 23.9 & 18.4 &  6.0 &  9.6 & 19.8 \\         ZICL mT0 &   87.3 & 90.5 & 72.7 & 90.8 & 93.0 & 89.5 \\\bottomrule\end{tabular}
     \caption{Model performance on \textsc{Amazon Reviews Multi}. 
     We report the average of the three few-shot samples. 
    }
    \label{tab:amazon_review_result}
\end{table*}

\begin{table*}[t!]
    \centering
    \footnotesize
\begin{tabular}{l|rrrrrr}\toprule      Transfer + Model &  Macro &   as &   bn &  brx &   gu &   hi \\\midrule        Target FT &   58.2 & 61.4 & 55.8 & 62.6 & 56.7 & 64.1 \\       English FT &   87.4 & 85.0 & 87.4 & 89.4 & 88.4 & 91.6 \\English Target FT &   88.4 & 84.6 & 90.2 & 90.6 & 89.7 & 93.0 \\       EICL BLOOM &   87.2 & 83.7 & 87.6 & 91.2 & 86.1 & 92.0 \\         EICL mT5 &   49.8 & 49.8 & 49.8 & 49.8 & 49.8 & 49.8 \\      EICL BLOOMZ &   93.0 & 89.6 & 94.2 & 94.9 & 93.1 & 95.6 \\         EICL mT0 &   79.9 & 73.6 & 88.4 & 81.3 & 80.2 & 81.1 \\     EICL ChatGPT &   89.3 & --& 91.8 & --& --& --\\       TICL BLOOM &   86.5 & 83.1 & 86.7 & 91.2 & 84.1 & 92.6 \\         TICL mT5 &   49.8 & 49.8 & 49.8 & 49.8 & 49.8 & 49.8 \\      TICL BLOOMZ &   64.5 & 67.0 & 61.2 & 94.9 & 52.8 & 56.5 \\         TICL mT0 &   69.0 & 87.4 & 82.9 & 50.1 & 78.2 & 68.3 \\     TICL ChatGPT &   89.7 & --& 92.6 & --& --& --\\       ZICL BLOOM &   49.7 & 49.8 & 49.8 & 49.8 & 49.8 & 49.8 \\         ZICL mT5 &   26.5 & 24.4 & 24.4 & 24.8 & 26.0 & 26.1 \\      ZICL BLOOMZ &  64.5 & 67.0 & 61.2 & 94.9 & 52.8 & 56.5 \\         ZICL mT0 &   93.2 & 90.5 & 93.7 & 94.3 & 92.2 & 95.3 \\\bottomrule\end{tabular}
\begin{tabular}{l|rrrrrrrrr}\toprule      Transfer + Model &   kn &  mai &   ml &   mr &   or &   pa &   ta &   te &   ur \\\midrule        Target FT & 59.5 & 62.6 & 45.8 & 60.4 & 62.7 & 48.9 & 57.8 & 55.0 & 60.8 \\       English FT & 88.4 & 89.4 & 86.9 & 86.1 & 77.2 & 90.4 & 87.0 & 86.7 & 90.3 \\English Target FT & 89.6 & 90.6 & 86.4 & 86.2 & 77.9 & 91.6 & 87.4 & 88.5 & 91.1 \\       EICL BLOOM & 83.0 & 91.2 & 85.8 & 88.9 & 85.8 & 89.0 & 85.0 & 86.0 & 85.1 \\         EICL mT5 & 49.8 & 49.8 & 49.8 & 49.8 & 49.8 & 49.8 & 49.8 & 49.8 & 49.8 \\      EICL BLOOMZ & 92.7 & 94.9 & 91.8 & 92.4 & 93.8 & 94.2 & 90.6 & 90.5 & 93.5 \\         EICL mT0 & 74.8 & 71.6 & 83.2 & 81.6 & 78.3 & 88.1 & 86.7 & 78.0 & 71.7 \\     EICL ChatGPT & --& --& --& --& --& --& 82.3 & --& 93.9 \\       TICL BLOOM & 81.8 & 91.2 & 84.0 & 88.2 & 85.0 & 88.2 & 85.3 & 85.1 & 84.1 \\         TICL mT5 & 49.8 & 49.8 & 49.8 & 49.8 & 49.8 & 49.8 & 49.8 & 49.8 & 49.8 \\      TICL BLOOMZ & 49.7 & 94.9 & 66.3 & 58.3 & 59.2 & 57.3 & 68.2 & 50.3 & 66.9 \\         TICL mT0 & 72.1 & 49.7 & 84.4 & 79.7 & 66.1 & 68.8 & 55.3 & 58.7 & 64.9 \\     TICL ChatGPT & --& --& --& --& --& --& 83.9 & --& 92.4 \\       ZICL BLOOM & 49.8 & 49.8 & 49.3 & 49.8 & 49.8 & 49.8 & 49.6 & 49.8 & 48.7 \\         ZICL mT5 & 26.8 & 24.8 & 29.0 & 20.7 & 22.4 & 32.4 & 25.4 & 28.9 & 34.5 \\      ZICL BLOOMZ & 26.8 & 24.8 & 29.0 & 20.7 & 22.4 & 32.4 & 25.4 & 28.9 & 34.5 \\         ZICL mT0 & 93.5 & 94.3 & 92.0 & 92.8 & 91.2 & 95.2 & 92.3 & 92.9 & 94.6 \\\bottomrule\end{tabular}
     \caption{Model performance on \textsc{Indic Sentiment}. 
     We report the average of the three few-shot samples. 
    }
    \label{tab:indic_results}
\end{table*}

\subsection{Commonsense}

\paragraph{\textsc{XCOPA}.}
The experimental results on \textsc{XCOPA} are available in Table~\ref{tab:xcopa_results}. 
On \textsc{XCOPA}, ChatGPT and mT0 (\textsc{Z EICL}) yield high performance across languages. ChatGPT achieves particularly notable performance in Italian (91.2\%).
On the other hand, all of the fine-tuning-based methods struggle, as the small size of the source datasets in English. 
This result indicates that for a task that lacks a large-scale training dataset even in high-resource languages, LLMs using in-context learning may often result in higher performance.  
We observed that mT0 \textsc{English FT} faces difficulties when applied to \textsc{XCOPA}. 
This could be attributed to the limited size of the \textsc{XCOPA} English set, which might not provide enough data for a smaller mT5-base model to acquire comprehensive task knowledge.

\begin{table*}[t!]
    \centering
    \footnotesize
\begin{tabular}{lrrrrrrrrrrrr}\toprule       Transfer + Model &  Macro &   et &   ht &   it &   id &   qu &   sw &   zh &   ta &   th &   tr &   vi \\\midrule        Target FT &   46.7 & 50.0 & 50.1 & 48.3 & 50.5 & 50.4 & 32.5 & 49.8 & 49.3 & 49.4 & 33.9 & 50.0 \\       English FT &   0.0 & 0.0& 0.0& 0.0& 0.0& 0.0& 0.0& 0.0& 0.0& 0.0& 0.0& 0.0 \\English Target FT &   49.9 & 50.3 & 49.9 & 49.6 & 49.2 & 50.5 & 50.4 & 50.4 & 49.2 & 50.7 & 49.5 & 49.4 \\       EICL BLOOM &   50.0 & 51.5 & 49.0 & 49.9 & 50.0 & 50.6 & 50.0 & 50.1 & 49.5 & 50.0 & 49.9 & 50.0 \\         EICL mT5 &   50.0 & 50.0 & 49.9 & 50.7 & 50.0 & 49.5 & 49.8 & 49.9 & 50.7 & 50.0 & 50.0 & 50.0 \\      EICL BLOOMZ &   50.5 & 50.7 & 51.2 & 50.9 & 50.0 & 52.7 & 49.9 & 50.0 & 50.1 & 49.8 & 49.8 & 50.0 \\         EICL mT0 &   57.1 & 60.7 & 60.6 & 53.4 & 59.8 & 50.0 & 61.6 & 64.1 & 51.9 & 54.1 & 54.1 & 58.1 \\     EICL ChatGPT &   76.7 & 87.6 & --& 91.2 & --& --& --& --& 54.6 & 62.6 & 87.4 & --\\       TICL BLOOM &   50.1 & 49.8 & 50.4 & 50.4 & 49.0 & 48.8 & 52.2 & 50.6 & 49.6 & 50.0 & 49.8 & 50.2 \\         TICL mT5 &   50.0 & 49.9 & 50.0 & 49.9 & 50.0 & 50.0 & 49.9 & 50.0 & 50.0 & 50.0 & 49.5 & 50.9 \\      TICL BLOOMZ &   50.5 & 45.6 & 50.8 & 50.4 & 53.4 & 47.4 & 49.8 & 51.8 & 53.2 & 50.0 & 49.4 & 53.4 \\         TICL mT0 &   52.8 & 50.4 & 51.9 & 51.0 & 51.9 & 50.6 & 53.7 & 50.5 & 50.1 & 50.6 & 54.3 & 65.5 \\     TICL ChatGPT &   74.4 & 89.2 & --& 91.6 & --& --& --& --& 49.5 & 55.7 & 86.2 & --\\       ZICL BLOOM &   50.9 & 51.8 & 48.8 & 51.2 & 51.4 & 50.6 & 51.2 & 53.6 & 52.4 & 48.2 & 49.8 & 50.6 \\         ZICL mT5 &   50.1 & 49.8 & 50.4 & 50.4 & 49.0 & 48.8 & 52.2 & 50.6 & 49.6 & 50.0 & 49.8 & 50.2 \\      ZICL BLOOMZ &   50.1 & 48.6 & 50.2 & 52.4 & 47.4 & 50.8 & 45.2 & 46.8 & 54.8 & 50.6 & 52.8 & 51.0 \\         ZICL mT0 &   64.1 & 64.0 & 62.2 & 66.2 & 70.0 & 48.8 & 66.2 & 71.8 & 61.0 & 63.0 & 65.0 & 67.2 \\\bottomrule\end{tabular}
     \caption{Model performance on \textsc{XCOPA}. 
     We report the average of the three few-shot samples. 
    }
    \label{tab:xcopa_results}
\end{table*}
\paragraph{\textsc{XWinograd}.}
The experimental results on \textsc{XWinograd} are available in Table~\ref{tab:xwinograd}. 
Similar to \textsc{XCOPA}, on \textsc{XWinograd}, fine-tuning-based methods often struggle, while in-context learning with competitive LLMs yields strong performance. 

\begin{table*}[t!]
    \centering
    \footnotesize
\begin{tabular}{lrrrrr}\toprule       Transfer + Model &  Macro &   jp &   pt &   ru &   zh \\\midrule        Target FT &   50.0 & 48.4 & 50.3 & 49.9 & 51.4 \\       English FT &   0.0 & 0.0& 0.0& 0.0& 0.0\\English Target FT &   49.0 & 48.4 & 48.4 & 48.8 & 50.6 \\       EICL BLOOM &   50.8 & 49.6 & 48.0 & 54.0 & 51.5 \\         EICL mT5 &   49.2 & 48.4 & 49.5 & 47.4 & 51.3 \\      EICL BLOOMZ &   52.1 & 52.6 & 50.3 & 55.3 & 50.1 \\         EICL mT0 &   59.6 & 62.2 & 57.7 & 53.2 & 65.2 \\     EICL ChatGPT &   73.3 & --& 74.1 & 72.5 & --\\       TICL BLOOM &   51.7 & 52.2 & 50.2 & 54.3 & 50.1 \\         TICL mT5 &   47.3 & 48.4 & 46.2 & 44.4 & 50.3 \\      TICL BLOOMZ &   53.1 & 52.7 & 54.5 & 55.3 & 50.0 \\         TICL mT0 &   57.9 & 54.9 & 57.2 & 62.9 & 56.5 \\     TICL ChatGPT &   71.6 & --& 70.4 & 72.8 & --\\       ZICL BLOOM &   53.7 & 51.9 & 54.4 & 56.7 & 51.9 \\         ZICL mT5 &   46.4 & 47.4 & 48.5 & 45.7 & 44.2 \\      ZICL BLOOMZ &   50.9 & 51.9 & 51.9 & 50.2 & 49.6 \\         ZICL mT0 &   64.5 & 68.7 & 59.8 & 62.2 & 67.3 \\\bottomrule\end{tabular}
     \caption{Model performance on \textsc{XWinograd}
     We report the average of the three few-shot samples. 
    }
    \label{tab:xwinograd}
\end{table*}

\subsection{Question Answering}
\textsc{TyDiQA} experimental results are available in Table~\ref{tab:tydiqa}. 
Both the fine-tuning and ICL methods exhibit commendable performance on this particular task. It is intriguing to note that both mT0 and BLOOMZ demonstrate relatively lower efficacy in Korean, Finnish, and Russian. This can be attributed to the fact that these languages were not included during the pretraining phase.

\begin{table*}[t!]
    \centering
    \footnotesize
\begin{tabular}{l|rrrrrrrrr}\toprule      Transfer + Model &  Macro &   ar &   be &   fi &   id &   sw &   ko &   ru &   te \\\midrule        Target FT &    8.3 &  8.1 &  6.1 &  9.1 &  6.4 &  5.5 &  7.5 &  9.2 & 14.7 \\       English FT &   62.9 & 61.0 & 63.2 & 65.3 & 69.2 & 67.9 & 57.1 & 56.3 & 63.5 \\English Target FT &   66.7 & 65.9 & 68.0 & 63.6 & 70.0 & 69.3 & 60.6 & 65.1 & 70.7 \\       EICL BLOOM &   39.2 & 43.8 & 58.2 & 20.6 & 47.0 & 57.5 & 23.2 & 32.7 & 30.4 \\         EICL mT5 &    0.3 &  0.7 &  0.1 &  0.4 &  0.2 &  0.3 & 0.0 &  0.3 & 0.0 \\      EICL BLOOMZ &   44.5 & 45.3 & 67.7 & 18.9 & 61.0 & 73.7 & 12.4 & 19.6 & 57.6 \\         EICL mT0 &   69.0 & 54.0 & 75.8 & 68.9 & 68.8 & 75.5 & 68.1 & 53.7 & 86.7 \\     EICL ChatGPT &   70.8 & --& 58.9 & --& 76.5 & 77.0 & --& --& --\\       TICL BLOOM &    7.0 & 13.2 & 11.9 &  1.7 & 19.1 &  4.5 &  0.7 &  1.3 &  3.7 \\         TICL mT5 &    0.2 &  0.4 &  0.1 &  0.2 &  0.6 &  0.2 & --&  0.3 & --\\      TICL BLOOMZ &   43.7 & 44.7 & 63.7 & 17.5 & 60.3 & 71.5 & 12.1 & 20.3 & 59.3 \\         TICL mT0 &   70.8 & 58.7 & 75.8 & 66.9 & 72.1 & 78.3 & 72.1 & 65.9 & 76.6 \\     TICL ChatGPT &   66.7 & --& 46.0 & --& 76.7 & 77.4 & --& --& --\\       ZICL BLOOM &    2.0 &  2.2 &  1.1 &  3.1 &  3.2 &  2.3 &  1.0 &  1.5 &  1.7 \\         ZICL mT5 &   65.2 & 80.0 & 86.3 &  7.3 & 81.3 & 82.0 & 44.7 & 55.0 & 85.1 \\      ZICL BLOOMZ &   65.2 & 80.0 & 86.3 &  7.3 & 81.3 & 82.0 & 44.7 & 55.0 & 85.1 \\         ZICL mT0 &   75.2 & 71.8 & 84.4 & 67.3 & 77.3 & 78.6 & 68.3 & 65.0 & 88.9 \\\bottomrule\end{tabular}
     \caption{Model performance on \textsc{TyDiQA}.
     We report the average of the three few-shot samples. 
    }
    \label{tab:tydiqa}
\end{table*}

\subsection{Named Entity Recognition}
\paragraph{\textsc{WikiANN}.}

Table~\ref{tab:wikiann} contains the results for \textsc{WikiANN}. We specifically present the few-shot results since we discovered that zero-shot baselines consistently exhibit extremely poor performance, often close to zero, primarily because generating the answer in the precise output format proves to be challenging.

It's important to acknowledge that the \data-Light \textsc{WikiANN} subset comprises languages that are relatively high-resource, which could potentially lead to an overestimation of ChatGPT's performance. When comparing the best fine-tuning method with ChatGPT in the \data-light languages, they generally perform competitively, with the exception of Finnish.

\begin{table*}[t!]
    \centering
    \footnotesize
\begin{tabular}{l|rrrrrrrrrr}\toprule      Transfer + Model &  Macro &   ta &   fr &   it &   ja &   vi &   be &   gu &   et &   th \\\midrule  Target FT &   43.7 & 0.2 & 59.0 & 55.5 & 43.9 & 58.3 & 63.5 & 26.0 & 54.4 & 23.7 \\       English FT &   52.2 & 0.8 & 78.2 & 79.4 & 56.1 & 80.5 & 73.9 & 24.0 & 60.5 & 10.7 \\English Target FT &   56.6 & 0.8 & 78.1 & 76.8 & 55.7 & 75.9 & 76.8 & 37.0 & 76.0 & 25.6 \\       EICL BLOOM &   32.8 & 0.6 & 51.6 & 51.0 & 22.1 & 53.8 & 25.6 & 22.3 & 37.0 &  1.7 \\         EICL mT5 &    1.6 &0.0& 0.0& 0.0& 0.0& 0.0&  3.3 &  0.3 & 0.0& 0.0\\      EICL BLOOMZ &   22.4 & 0.5 & 37.1 & 43.4 & 15.6 & 36.8 & 15.4 & 13.0 & 29.6 &  0.3 \\         EICL mT0 &   15.8 & 0.1 & 13.8 & 13.0 &  9.1 & 22.9 & 11.0 &  6.0 & 24.1 &  1.4 \\     EICL ChatGPT &   77.6 &--& --& 81.8 & --& --& 78.2 & --& 78.2 & --\\       TICL BLOOM &   32.8 & 0.7 & 52.5 & 50.2 & 20.8 & 53.5 & 24.4 & 24.0 & 34.0 &  1.0 \\         TICL mT5 &    0.3 &0.0& 0.0&  0.1 & 0.0&  0.1 &  0.2 &  1.3 & 0.0&  1.7 \\      TICL BLOOMZ &   20.7 & 0.6 & 37.3 & 39.8 & 15.0 & 32.1 & 13.5 &  8.7 & 25.1 &  0.2 \\         TICL mT0 &   15.8 & 0.1 & 13.8 & 13.0 &  9.1 & 22.9 & 11.0 &  6.0 & 24.1 &  1.4 \\     TICL ChatGPT &   76.8 &--& --& 82.3 & --& --& 78.4 & --& 76.9 & --\\
\bottomrule\end{tabular}
\begin{tabular}{l|rrrrrrrrrrr}\toprule      Transfer + Model &   or &   kn &   fi &   gn &   ru &   el &   ur &   es &   hi &  te &   as \\\midrule        Target FT & 36.5 & 12.5 & 55.5 & 60.3 & 50.1 & 59.0 & 68.4 & 54.9 & 42.4 & 7.0 & 25.3 \\       English FT & 35.5 & 11.0 & 64.2 & 71.0 & 60.4 & 73.4 & 79.6 & 75.7 & 47.9 & 6.6 & 26.0 \\English Target FT & 40.0 & 22.5 & 74.8 & 68.0 & 67.8 & 74.4 & 79.1 & 78.3 & 53.7 & 9.5 & 28.3 \\       EICL BLOOM & 22.0 &  6.0 & 39.5 & 47.3 & 26.1 & 20.4 & 70.7 & 55.2 & 40.2 & 5.6 & 22.7 \\         EICL mT5 & 0.0&  1.3 & 0.0& 0.0& 0.0& 0.0& 10.1 & 0.0& 10.0 &0.0&  0.7 \\      EICL BLOOMZ & 10.0 &  5.7 & 31.8 & 28.0 & 19.7 & 15.8 & 41.7 & 37.5 & 30.9 & 4.2 & 16.0 \\         EICL mT0 & 16.3 &  3.3 & 15.2 & 24.3 & 15.1 & 12.8 & 47.1 & 20.3 & 18.7 & 3.3 & 10.0 \\     EICL ChatGPT & --& --& 81.5 & --& --& 72.4 & --& --& --&--& --\\       TICL BLOOM & 25.3 &  6.7 & 37.6 & 49.0 & 26.2 & 19.7 & 71.7 & 55.6 & 39.9 & 5.3 & 24.0 \\         TICL mT5 &  0.3 & 0.0& 0.0& 0.0& 0.0& 0.0& 0.0& 0.0&  1.8 &0.0&  1.0 \\      TICL BLOOMZ &  6.5 &  4.0 & 26.5 & 24.7 & 17.4 & 13.0 & 47.3 & 41.1 & 26.5 & 3.8 & 13.0 \\         TICL mT0 & 16.3 &  3.3 & 15.2 & 24.3 & 15.1 & 12.8 & 47.1 & 20.3 & 18.7 & 3.3 & 10.0 \\     TICL ChatGPT & --& --& 81.9 & --& --& 69.3 & --& --& --&--& --\\ 
\bottomrule\end{tabular}
\\
\begin{tabular}{l|rrrrrrrrrrrrrr}\toprule      Transfer + Model &   sw &   pa &   bg &   ml &   tr &   fa &   id &   ko &   mr &   de &   ar &   bn &   zh \\\midrule        Target FT & 57.5 & 29.7 & 54.2 & 19.7 & 55.4 & 48.0 & 64.2 & 36.1 & 34.8 & 51.2 & 40.6 & 43.0 & 49.9 \\       English FT & 61.0 & 35.5 & 67.0 & 21.4 & 64.5 & 60.5 & 81.6 & 36.2 & 36.6 & 75.1 & 52.9 & 48.7 & 66.6 \\English Target FT & 75.3 & 42.3 & 67.1 & 24.5 & 79.5 & 57.6 & 80.7 & 57.7 & 44.7 & 73.2 & 52.9 & 47.7 & 65.2 \\       EICL BLOOM & 60.3 & 26.3 & 30.9 & 14.0 & 39.4 & 28.6 & 61.2 & 12.0 & 28.4 & 41.7 & 43.9 & 34.9 & 38.7 \\         EICL mT5 & 0.0&  0.7 & 0.0& 0.0& 0.0& 0.0&  0.3 & 0.0&  0.4 &  6.7 & 16.7 &  3.7 & 0.0\\      EICL BLOOMZ & 34.9 & 15.0 & 22.7 &  5.0 & 34.6 & 14.7 & 31.7 &  9.8 & 22.6 & 26.4 & 21.0 & 36.0 & 31.3 \\         EICL mT0 & 24.3 & 10.0 & 14.7 &  5.0 & 20.2 & 21.4 & 23.4 & 11.2 & 12.3 & 15.7 & 23.0 & 23.9 & 27.7 \\     EICL ChatGPT & --& --& 73.3 & --& --& --& --& --& --& --& --& --& --\\       TICL BLOOM & 58.8 & 26.7 & 29.6 & 14.4 & 39.6 & 27.8 & 61.4 & 10.6 & 27.9 & 43.3 & 44.6 & 36.8 & 38.3 \\         TICL mT5 &  0.4 & & & & & & & &  0.5 &  0.1 &  0.4 &  0.3 & \\      TICL BLOOMZ & 26.8 & 14.0 & 19.7 &  4.2 & 31.3 & 14.7 & 35.2 &  8.1 & 20.4 & 22.4 & 23.6 & 36.2 & 31.0 \\         TICL mT0 & 24.3 & 10.0 & 14.7 &  5.0 & 20.2 & 21.4 & 23.4 & 11.2 & 12.3 & 15.7 & 23.0 & 23.9 & 27.7 \\     TICL ChatGPT & --& --& 72.0 & --& --& --& --& --& --& --& --& --& --\\      
\bottomrule\end{tabular}
     \caption{Model performance on \textsc{WikiANN}. 
     We report the average of the three few-shot samples. 
    }
    \label{tab:wikiann}
\end{table*}

\paragraph{\textsc{MasakhaNER}.}
Results on \textsc{MasakhaNER} are available at Table~\ref{tab:masakhaNER}. 
In this benchmark, all ICL methods, including ChatGPT, encounter difficulties, whereas \textsc{Target FT} and \textsc{Eng.+Tgt. FT} consistently demonstrates strong performance across various languages. Notably, by incorporating an additional 32 training examples, \textsc{Eng.+Tgt. FT} achieves a significant 34\% improvement in performance for Hausa. These remarkable enhancements underscore the effectiveness of fine-tuning a specialized model on a limited set of training samples in target languages.

\begin{table*}[t!]
    \centering
    \footnotesize
\begin{tabular}{l|rrrrrrrrrr}\toprule      Transfer + Model &  Macro &  amh &  hau &  ibo &  kin &  luo &  pcm &  swa &  wol &  yor \\\midrule  Target FT &   17.4 & 13.6 & 31.5 & 28.6 & 12.8 & 14.2 & 11.1 & 26.4 &  8.7 &  9.9 \\       English FT &    9.4 &  6.2 & 11.0 & 14.8 & 10.5 & 10.5 &  8.7 & 10.4 &  3.8 &  8.3 \\English Target FT &   30.5 & 27.0 & 44.7 & 44.3 & 26.8 & 26.0 & 23.7 & 40.6 & 18.8 & 22.4 \\       EICL BLOOM &   17.2 &  3.4 & 23.8 & 27.4 & 18.5 & 11.6 & 15.2 & 24.9 & 16.3 & 13.9 \\         EICL mT5 &    1.5 & 0.0& 13.3 & 0.0& 0.0&  0.4 & 0.0& 0.0& 0.0& 0.0\\      EICL BLOOMZ &   14.9 &  0.2 & 11.3 & 28.4 & 14.3 &  4.6 & 12.4 & 24.4 & 17.7 & 21.0 \\         EICL mT0 &    1.3 & 0.0&  1.7 &  0.8 &  4.9 &  1.2 & 0.0&  2.2 & 0.0&  0.8 \\     EICL ChatGPT &   13.2 & --& --& --& --& --& --& --& --& 13.2 \\       TICL BLOOM &   17.2 &  3.4 & 23.8 & 27.4 & 18.5 & 11.6 & 15.2 & 24.9 & 16.3 & 13.9 \\         TICL mT5 &    0.2 & 0.0&  1.6 & 0.0& 0.0&  0.4 & 0.0& 0.0& 0.0& 0.0\\      TICL BLOOMZ &   14.9 &  0.2 & 11.3 & 28.4 & 14.3 &  4.6 & 12.4 & 24.4 & 17.7 & 21.0 \\         TICL mT0 &    1.3 & 0.0&  1.7 &  0.8 &  4.9 &  1.2 & 0.0&  2.2 & 0.0&  0.8 \\     TICL ChatGPT &   12.8 & --& --& --& --& --& --& --& --& 12.8 \\ 
\bottomrule\end{tabular}
     \caption{Model performance on \textsc{MasakhaNER}.
     We report the average of the three few-shot samples. 
    }
    \label{tab:masakhaNER}
\end{table*}

\subsection{Generation }
\paragraph{\textsc{TyDiQA-QG}.}
The experimental results for \textsc{TyDiQA-QG} are available in Table~\ref{tab:tydiqa_qg}. 
On this task, ChatGPT and mT0 \textsc{English ICL} show superior performance than smaller fine-tuned models, demonstrating their competitiveness in generating fluent text in target languages. 

\begin{table*}[t!]
    \centering
    \footnotesize
\begin{tabular}{l|rrrrrrrrr}\toprule      Transfer + Model &  Macro &   ar &   be &   fi &   id &   sw &   ko &   ru &   te \\\midrule        Target FT &    3.4 &  2.7 &  4.1 &  2.5 &  4.4 &  3.2 &  2.8 &  2.1 &  5.8 \\       English FT &    4.2 &  2.1 &  3.5 &  5.1 &  6.2 &  5.1 &  3.0 &  4.7 &  4.2 \\English Target FT &   12.2 & 11.5 &  7.3 & 15.8 & 14.1 & 13.1 &  7.9 &  8.9 & 18.8 \\       EICL BLOOM &   11.6 & 18.3 & 10.4 & 10.8 & 16.1 & 15.2 &  1.3 &  3.7 & 17.4 \\         EICL mT5 &   0.0& 0.0& 0.0& 0.0& 0.0& 0.0& 0.0& 0.0&  0.1 \\      EICL BLOOMZ &   13.9 & 19.5 & 14.2 &  7.8 & 23.6 & 23.1 &  0.7 &  2.1 & 20.3 \\         EICL mT0 &   15.3 & 25.8 & 10.3 &  3.7 & 19.6 & 12.3 &  4.1 &  6.2 & 40.1 \\     EICL ChatGPT &   17.8 & 30.6 & --& 28.2 & --& --&  0.7 &  2.6 & 26.9 \\       TICL BLOOM &   12.8 & 18.1 &  9.6 & 10.0 & 15.7 & 14.9 &  7.7 &  9.2 & 16.8 \\         TICL mT5 &   0.0& 0.0& 0.0& 0.0& 0.0& 0.0& 0.0& 0.0& 0.0\\      TICL BLOOMZ &   12.0 & 16.0 & 10.7 &  5.0 & 20.0 & 21.1 &  1.9 &  5.2 & 15.9 \\         TICL mT0 &   14.6 & 17.7 &  9.1 &  6.6 & 18.3 & 12.0 &  5.1 &  8.5 & 39.3 \\     TICL ChatGPT &   19.2 & 24.0 & --& 27.5 & --& --& 14.8 & 17.6 & 12.2 \\       ZICL BLOOM &    0.1 &  0.1 &  0.1 & 0.0&  0.1 &  0.1 &  0.1 &  0.1 & 0.0\\         ZICL mT5 &   16.5 & 30.6 & 15.5 &  5.2 & 24.5 & 21.8 &  3.0 &  4.6 & 26.8 \\      ZICL BLOOMZ &    1.7 &  2.4 &  2.1 &  1.7 &  2.5 &  2.2 &  1.0 &  0.9 &  1.2 \\         ZICL mT0 &   10.3 &  4.9 & 13.7 &  3.5 & 12.3 &  5.4 &  1.9 &  2.0 & 39.1 \\\bottomrule\end{tabular}
     \caption{Model performance on \textsc{TyDiQA-QG}. 
     We report the average of the three few-shot samples. 
    }
    \label{tab:tydiqa_qg}
\end{table*}

\paragraph{\textsc{XLSum}.}
\textsc{XLSum} results are available in Table~\ref{tab:xlsum_results}. 
Despite strong generation capability, ChatGPT \textsc{English ICL} performance remains low. We found that when instructed in English, ChatGPT often generates summaries in English, not in the article language. We haven't observed such behaviors on other tasks or other LLMs. 
ChatGPT \textsc{Target ICL} shows large improvements from \textsc{English ICL}, which has not been observed in other tasks. When instructions in the target language are given, ChatGPT almost always generates a summary in the target language. 

Among non-instruction-tuned models, \textsc{Eng.+Tgt. FT} yields the highest average performance. 
It should be noted that mT0 and BLOOMZ are trained on \textsc{XLSum}. Nevertheless, their performance in some languages remains low.

\begin{table*}[t!]
    \centering
    \footnotesize
\begin{tabular}{l|rrrrr}\toprule      Transfer + Model &  Macro &  Tamil &  Vietnamese &  Swahili &  Indonesian \\\midrule        Target FT &    2.8 &    0.8 &        11.0 &      2.0 &         1.7 \\       English FT &    4.0 &    0.1 &        18.4 &      7.8 &         4.9 \\English Target FT &    8.4 &   10.9 &        24.7 &      8.8 &         7.8 \\       EICL BLOOM &    2.4 &    0.1 &         9.0 &      4.6 &         3.8 \\         EICL mT5 &    0.3 &   0.0&         1.7 &      0.4 &         0.2 \\      EICL BLOOMZ &    9.0 &   18.6 &        12.3 &      1.6 &         3.3 \\         EICL mT0 &    1.8 &   0.0&        10.4 &      5.3 &         1.0 \\     EICL ChatGPT &    5.4 &   --&        19.5 &     --&         4.9 \\       TICL BLOOM &    4.7 &   13.9 &        10.3 &      4.6 &         3.1 \\         TICL mT5 &    0.3 &   0.0&         1.7 &      0.3 &         0.3 \\      TICL BLOOMZ &   10.9 &    4.6 &        12.9 &      1.2 &        15.7 \\         TICL mT0 &    1.8 &  0.0&        10.4 &      5.3 &         1.0 \\     TICL ChatGPT &   11.4 &   --&        19.5 &     --&         7.2 \\       ZICL BLOOM &    4.1 &    0.1 &        10.7 &      9.0 &         9.5 \\         ZICL mT5 &    1.3 &    0.5 &         4.8 &      1.1 &         0.7 \\      ZICL BLOOMZ &    4.3 &  0.0&        0.0&    0.0&         9.5 \\         ZICL mT0 &    8.5 &    1.1 &        26.9 &     18.3 &        16.8 \\\bottomrule\end{tabular}
\begin{tabular}{l|rrrrrrrr}\toprule      Transfer + Model &  Turkish &  Japanese &  Thai &  Bengali &  Arabic &  Spanish &  Persian &  Chinese \\\midrule        Target FT &   1.1 &       6.5 &   6.5 &    0.0&   0.0&      1.5 &    0.0&                 2.2 \\       English FT &      8.0 &       0.7 &   0.9 &    0.0&   0.0&      5.7 &    0.0&                 1.2 \\English Target FT &     12.1 &       2.8 &   8.5 &     0.0&     3.3 &     10.7 &     10.0 &                 1.5 \\       EICL BLOOM &      5.2 &       0.3 &   0.2 &     0.0&     0.1 &      3.7 &     0.0&                 1.1 \\         EICL mT5 &      0.4 &      0.0&  0.0&     0.0&    0.0&      0.4 &     0.0&                0.0\\      EICL BLOOMZ &      7.0 &       0.9 &  48.6 &     0.0&    0.0&      5.0 &     10.0 &                 0.4 \\         EICL mT0 &      1.0 &      0.0&  0.0&     0.0&    0.0&      4.0 &     0.0&                0.0\\     EICL ChatGPT &      2.4 &      --&  --&     --&    --&     --&     --&                --\\       TICL BLOOM &      5.2 &      14.1 &   0.5 &     0.0&    0.0&      3.6 &     0.0&                 1.2 \\         TICL mT5 &      0.5 &      0.0&  0.0&     0.0&    0.0&      0.4 &     0.0&                0.0\\      TICL BLOOMZ &      3.2 &      37.4 &  48.6 &     0.0&    0.0&      5.8 &     0.0&                 1.5 \\         TICL mT0 &      1.0 &      0.0&  0.0&     0.0&    0.0&      4.0 &     0.0&                0.0\\     TICL ChatGPT &     10.0 &      --&  --&     --&    --&     --&     20.1 &                --\\       ZICL BLOOM &      4.3 &       0.8 &   0.2 &     0.0&    0.0&      3.3 &     10.0 &                 1.6 \\         ZICL mT5 &      1.1 &       2.4 &   1.9 &     0.0&     0.1 &      0.7 &     0.0&                 1.9 \\      ZICL BLOOMZ &     0.0&      0.0&  0.0&     0.0&    0.0&      7.6 &      0.1 &                0.0\\         ZICL mT0 &     15.7 &       3.1 &   2.4 &     0.0&     0.1 &     12.4 &      0.2 &                 4.4 \\\bottomrule\end{tabular}
     \caption{Model performance on \textsc{XLSum}
    }
    \label{tab:xlsum_results}
\end{table*}

\section{More Analysis}

\subsection{Performance across Languages}
\begin{figure*}[t!]
\centering
\includegraphics[width=15.5cm]{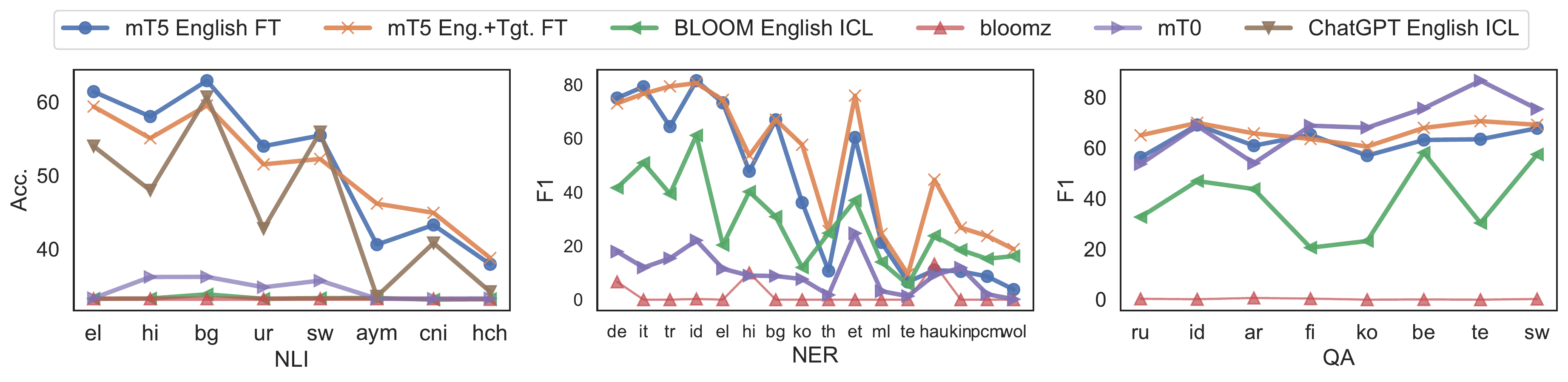}\caption{
\textbf{Model performance across three tasks, NLI, NER, and QA, displayed for various languages.} The languages are sorted based on token availability in mC4, with the left side representing high-resource languages. {All methods show performance deterioration in lower-resource languages (right side), with larger drops in \textsc{English-ICL} methods. Additional fine-tuning in target languages is more effective in less-represented languages. }
} \label{fig:performance_diff_more_methods}
\end{figure*}

Figure~\ref{fig:performance_diff_more_methods} shows performance across languages on the three tasks, NLI, NER, and QA, adding two more LLMs: BLOOMZ and mT0. 

\subsection{Variances of Different $k$-shots}
In Section~\ref{fig:demo_scaling}, we show that different sets of demonstrations can cause significant performance differences. 
We provide the full visualization results across different tasks. 

\subsection{Variances of the Varying Number of $k$}
\label{sec:more_varying_demo}
\begin{figure*}[t!]
\centering
\includegraphics[width=15.5cm]{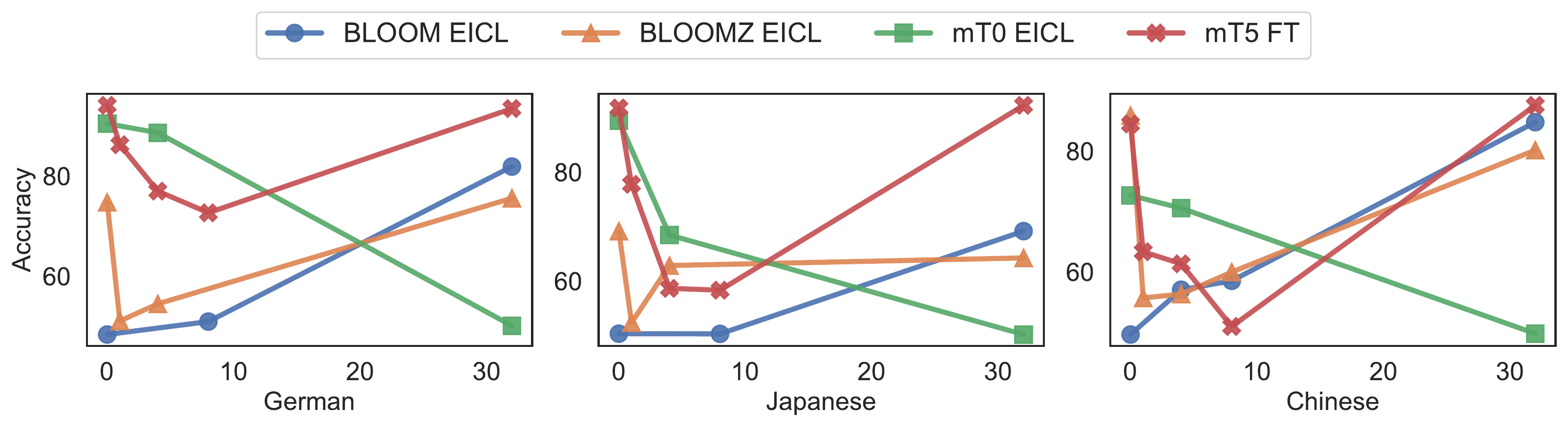}\caption{
\textbf{Effects of demonstrations on Amazon Review.} The $x$-axis indicates the number of training instances used during the transfer. 
} \label{fig:demo_scale_amazon}
\end{figure*}

\begin{figure*}[t!]
\centering
\includegraphics[width=15.5cm]{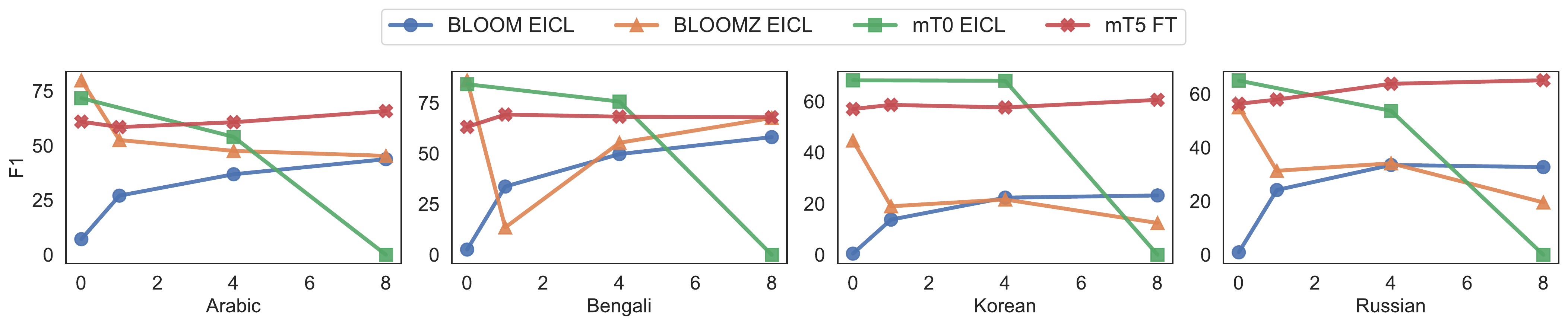}\caption{
\textbf{Effects of demonstrations on \textsc{TyDiQA}.} The $x$-axis indicates the number of training instances used during the transfer.
} \label{fig:demo_scale_tydiqa}
\end{figure*}

\begin{figure*}[t!]
\centering
\includegraphics[width=15.5cm]{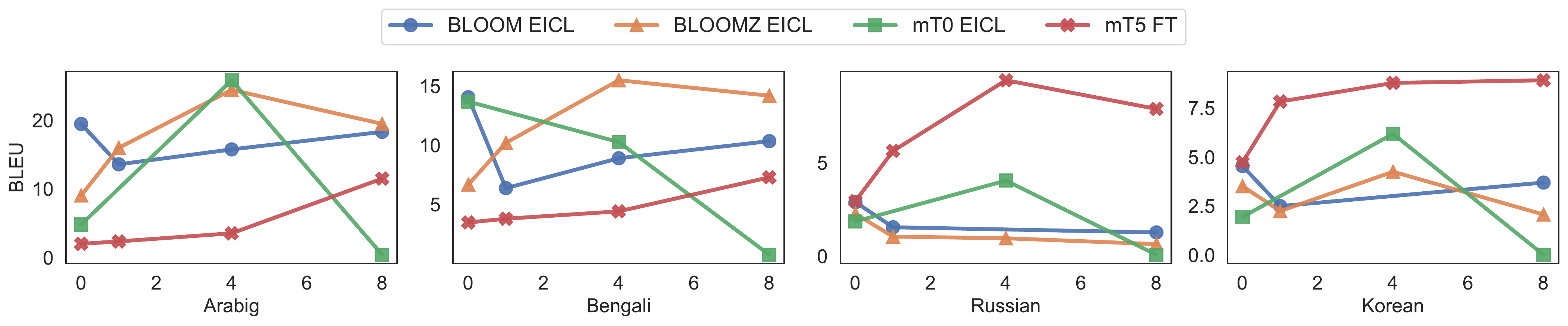}\caption{
\textbf{Effects of demonstrations on \textsc{TyDIQA-QG}.}  The $x$-axis indicates the number of training instances used during the transfer.
} \label{fig:demo_scale_tydiqg}
\end{figure*}

\begin{figure*}[t!]
\centering
\includegraphics[width=15.5cm]{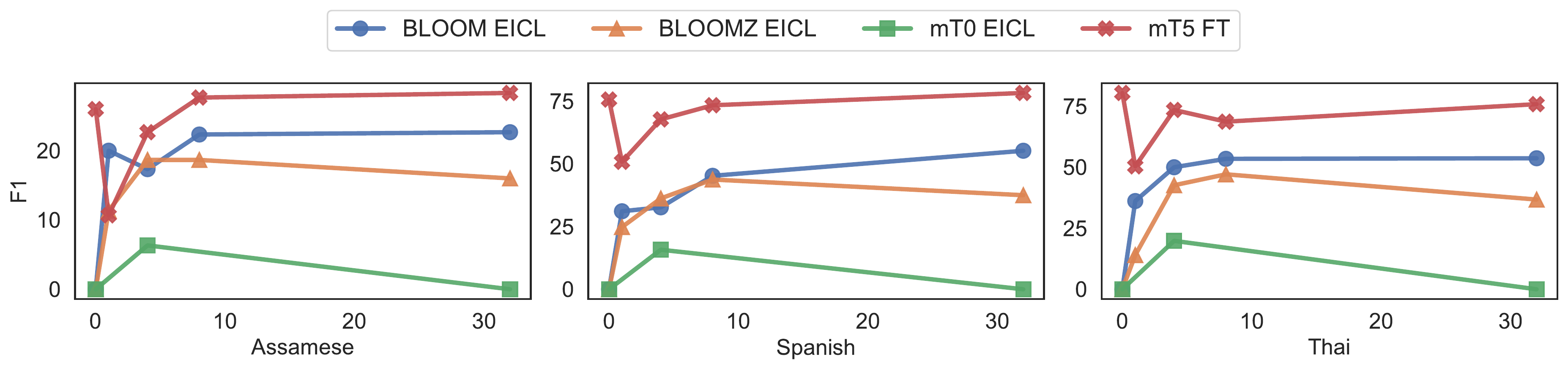}\caption{
\textbf{Effects of demonstrations on \textsc{WikiANN}.} The $x$-axis indicates the number of training instances used during the transfer.
} \label{fig:demo_scale_wikiann}
\end{figure*}

We provide the full experimental results with a different number of $k$.
We evaluate BLOOM  \textsc{English ICL}, BLOOMZ  \textsc{English ICL} and mT5- \textsc{Eng.+Tgt. Fine-tuning} and mT0 \textsc{English ICL} experimental results on \textsc{Amazon Review}, \textsc{TyDiQA}, \textsc{TyDiQA-AG}, \textsc{WikiANN}, and in Figures~\ref{fig:demo_scale_amazon}, \ref{fig:demo_scale_tydiqa}, \ref{fig:demo_scale_tydiqg} and \ref{fig:demo_scale_wikiann}, respectively. 

\paragraph{\textsc{Amazon Review. }}
On \textsc{Amazon Review}, All models except for BLOOM (pretraining only) show competitive zero-shot performance. BLOOM \textsc{English ICL} benefits from few-shot demonstrations while mT0 \textsc{English ICL} exhibit performance deterioration as adding more demonstrations across languages. 

\paragraph{\textsc{TyDiQA.}}
Among \textsc{English ICL} baselines, mT0 shows strong performance up to four demonstrations, although their performance gets really low once more demonstrations are added. Similar deterioration happens in BLOOMZ. On the contrary, BLOOM performance improves as more shots are added. 
Despite using only 32 shots. 

\paragraph{\textsc{TyDiQA-QG.}}
Unlike in \textsc{Amazon Review} or \textsc{TyDiQA}, BLOOMZ \textsc{English ICL} shows performance improvements with more demonstrations in Arabic and Bengali, reaching the highest QG performance in Bengali with four demonstrations. On the contrary, both BLOOM and BLOOMZ show poor QG performance in Korean and Russian, possibly due to the lack of those languages during pretraining.  

\paragraph{\textsc{WikiANN.}}
On WikiANN, all of the models gain performance improvements by adding at least one demonstration, possibly due to the difficulty of learning the exact output format expected given the instruction only. As in other datasets, mT0 reaches its highest performance with four demonstrations. mT5 \textsc{Eng.+Tgt. FT} exhibits performance drops with one shot, possibly due to their overfit to the single example. 

\subsection{Variances of Different Instructions}

\begin{table*}[t!]
    \centering \footnotesize
    \begin{tabular}{p{5cm}
        R{.5cm}R{.5cm}R{.5cm}R{.5cm}
        R{.5cm}R{.6cm}R{.5cm}R{.5cm}
        R{.5cm}R{.5cm}R{.5cm}R{.5cm}
        }
\toprule
\multirow{3}{*}{Instruction} &
            \multicolumn{6}{c}{BLOOM} & \multicolumn{6}{c}{mT0} \\
            & \multicolumn{3}{c}{0-shot} & \multicolumn{3}{c}{4-shot}
            & \multicolumn{3}{c}{0-shot} & \multicolumn{3}{c}{4-shot}
            \\
            \cmidrule(lr){2-4} \cmidrule(lr){5-7} \cmidrule(lr){8-10} \cmidrule(lr){11-13}
            & fi & ru & sw & fi & ru & sw & fi & ru & sw & fi & ru & sw  \\
            \midrule
This task is about reading the given passage and constructing a question about the information present in the passage. Construct a question in such a way that (i) it is unambiguous, (ii) it is answered from the passage, (iii) its answer is unique (iv) its answer is a continuous text span from the paragraph. Avoid creating questions that (i) can be answered correctly without actually understanding the paragraph and (ii) uses the same words or phrases given in the passage. & 5.1 & 3.3 & 5.1 & 8.7 & 4.3 & \textbf{10.8} & 4.0 & 3.7 & 5.0 & 5.0 & 5.3 & 3.1\\\midrule
Could you generate a question in \texttt{lang} whose answer is as provided based on the following context? & 5.1 & 3.3 & 5.1 & 9.1 & 4.3 & 9.5 & 4.0 & 3.7 & 5.0 & 6.5 & 7.5 & 8.7\\\midrule
Could you generate a \texttt{lang} question whose answer is as provided based on the following context? & 5.1 & 3.3 & 5.1 & 9.2 & 4.3 & 9.3 & 4.0 & 3.7 & 5.0 & 6.6 & 7.4 & 8.4 \\\midrule
Generate a \texttt{lang} question whose answer is as provided based on the following context. & 5.1 & 3.3 & 5.1 & \textbf{9.3} & \textbf{4.4} & 9.1 & 4.0 & 3.7 & 5.0 & \textbf{7.1} & \textbf{7.7} & \textbf{9.0}\\\midrule
In this task, you are given a review from Amazon. Your task is to generate a rating for the product on a scale of 1-5 based on the review. The rating means -2: extremely poor, 1: poor, 0: neutral, 1: good, 2: extremely good, 3: mixed and -3: no sentiment expressed.& 5.1 & 3.3 & 5.1 & 9.2 & \textbf{4.4} & 10.4 & 4.0 & 3.7 & 5.0 & 0.4 & 0.1 & 0.4  \\
\bottomrule
 \end{tabular}
    \caption{The performance (in BLEU score) for different instructions for \textsc{TyDiQA-QG}. \texttt{lang} stands for the full language name, e.g., Arabic. We modify the instructions from \data, PromptSource~\citep{bach2022promptsource}, and SuperNaturalInstructions~\citep{wang-etal-2022-super}.
    }
    \label{tab:instructions_tydiqaqg_ablation}
\end{table*}

We investigate the effectiveness of different English instructions on question generation tasks for \textsc{TyDiQA} in 0 and 4-shot setting using mT0 and BLOOM as base models {in Table \ref{tab:instructions_tydiqaqg_ablation}}.
We compare four relevant instructions and one irrelevant instruction (an instruction for \textsc{Amazon Review}). 
In a zero-shot setting, instruction does not make much difference for both instruction-tuned and non-instruction-tuned models, since irrelevant instructions are sometimes better than the relevant prompt.

In a four-shot setting, whether the instruction is relevant does not make a huge difference for BLOOM, and we observed that random seeds impact the performance more, yet the performances do suffer a sharp loss if we are using irrelevant instruction in the instruction-tuned model. 
{We also discovered that different models might favor different instructions for different languages, for example, in Swahili, 4-shot BLOOM favors the first instruction, while mT0 favors the fourth instruction. }

\subsection{Qualitative Results for Generation Tasks}
Table~\ref{tab:xlsum_examples} shows some qualitative results of ChatGPT \textsc{English ICL} and \textsc{Target TCL} on \textsc{XLSum} and \textsc{TyDiQA}.  
Given English instruction, ChatGPT often generates summaries in English, rather than in the article language. 
On the other hand, such cross-lingual generation behaviors don't occur in QA tasks, and the model's predictions with \textsc{Target ICL} and \textsc{English ICL} exhibit high overlap with each other. 
We hypothesize that ChatGPT's this cross-lingual summarization behavior can be related to their private training corpus, and future work can further investigate this issue. 

\begin{table*}[ht!]
\center
\small
\begin{tabular}{ p{0.1\linewidth} p{0.4\linewidth}  p{0.45\linewidth}}
\toprule
Target &  Gold. & Pred.  \\  
\hline
\multirow{8}{\linewidth}{\textsc{XLSum}-Indonesian} & \multirow{8}{\linewidth}{Manajer Chelsea Jose Mourinho memberi waktu 48 jam untuk Wayne Rooney memutuskan apakah dia ingin meninggalkan Manchester United dan pindah ke Stamford Bridge. (Google Translate: Chelsea manager Jose Mourinho has given Wayne Rooney 48 hours to decide whether he wants to leave Manchester United and move to Stamford Bridge.)} & \multirow{3}{\linewidth}{\textsc{English ICL:} Manchester United striker Wayne Rooney has been given 48 hours to resolve speculation over his future, with Chelsea expressing interest in the player. \hl{[en]}} \\
  \\
   \\\\
   & & \multirow{3}{\linewidth}{\textsc{Target ICL:}Mourinho memberi ultimatum kepada Wayne Rooney untuk mengakhiri spekulasi kepindahannya dalam waktu 48 jam, sementara MU dikabarkan sedang mengincar Gareth Bale.  \hl{[id]}}
  \\\\\\
   \\\midrule
   \multirow{6}{\linewidth}{\textsc{XLSum}-Turkish} & \multirow{6}{\linewidth}{Suriye'nin Rakka vilayetine bağlı Tel Abyad kent merkezinde IŞİD ile YPG güçleri arasında şiddetli çatışmalar sürerken, Akçakale sınırı sığınmacılara yeniden açıldı. IŞİD'e karşı savaşan YPG güçleri de Tal Abyad'a girdi.} & \multirow{3}{\linewidth}{\textsc{English ICL:} YPG forces have cut off a vital ISIS supply route from Tel Abyad to Raqqa, and have taken control of two neighborhoods in Tel Abyad. \hl{[en]}} \\
   \\\\
&& \multirow{4}{\linewidth}{\textsc{Target ICL:} YPG\'nin ele geçirdiği, çatışmaların devam ettiği ve sivillerin tehlikede olduğu" bilgisini verdi. \hl{[tr]}} \\
\\\\\midrule
\multirow{5}{\linewidth}{\textsc{TyDiQA}-Swahili} & \multirow{5}{\linewidth}{'Sehemu ya chakula pamoja na wanga, protini na vitamin} & \multirow{2}{\linewidth}{\textsc{English ICL:} sehemu za chakula pamoja na wanga, protini na vitamini.\hl{[sw]}} \\
\\\\
&& \multirow{2}{\linewidth}{\textsc{Target ICL:} Sehemu za chakula pamoja na wanga, protini na vitamini. \hl{[sw]}} \\ \\
\bottomrule
\end{tabular}
\caption{ChatGPT outputs for \textsc{XLSum} and \textsc{TyDiQA} Output. \hl{[lang]} indicate the language codes of the output text languages. 
}
\label{tab:xlsum_examples}
\end{table*}